\documentclass[10pt,journal,compsoc]{IEEEtran}

\ifCLASSOPTIONcompsoc
  \usepackage[nocompress]{cite}
\else
  \usepackage{cite}
\fi

\ifCLASSINFOpdf
\else
\fi
\usepackage[pagebackref=true,breaklinks=true,letterpaper=true,colorlinks,bookmarks=false]{hyperref}
\usepackage{times}  
\usepackage{helvet}  
\usepackage{courier}  
\usepackage{url}  
\usepackage{graphicx}  
\usepackage[usenames, dvipsnames]{color}

\usepackage{amsmath}
\usepackage{amssymb}
\usepackage{multirow}
\usepackage{multicol}
\usepackage{setspace}
\usepackage{booktabs}

\usepackage{caption}
\usepackage{subcaption}

\usepackage{enumitem}

\usepackage{etoolbox}
\patchcmd{\thebibliography}
  {\settowidth}
  {\setlength{\itemsep}{0pt plus 0.1pt}\settowidth}
  {}{}
\apptocmd{\thebibliography}
  {\small}
  {}{}

\setlist{nolistsep}  

\begin{document}

\title{Operator-in-the-Loop Deep Sequential Multi-camera Feature Fusion for Person Re-identification}

\author{K.~L.~Navaneet,
        Ravi~Kiran~Sarvadevabhatla,~\IEEEmembership{Member,~IEEE},
        Shashank Shekhar,
        R.~Venkatesh~Babu,~\IEEEmembership{Senior Member,~IEEE},
        and~Anirban~Chakraborty$^*$\thanks{* Corresponding author},~\IEEEmembership{Member,~IEEE}
\thanks{The authors are with the Department of Computational and Data Sciences, Indian Institute of Science, Bangalore, India, 560012. \break E-mail: navaneetl@iisc.ac.in, ravika@gmail.com, shashankshek@iisc.ac.in, venky@iisc.ac.in, anirban@iisc.ac.in.}}

\IEEEtitleabstractindextext{%
\begin{abstract}
Given a target image as query, person re-identification systems retrieve a ranked list of candidate matches on a per-camera basis. In deployed systems, a human operator scans these lists and labels sighted targets by touch or mouse-based selection. However, classical re-id approaches generate per-camera lists independently. Therefore, target identifications by operator in a subset of cameras cannot be utilized to improve ranking of the target in remaining set of network cameras. To address this shortcoming, we propose a novel sequential multi-camera re-id approach. The proposed approach can accommodate human operator inputs and provides early gains via a monotonic improvement in target ranking. At the heart of our approach is a fusion function which operates on deep feature representations of query and candidate matches. We formulate an optimization procedure custom-designed to incrementally improve query representation. Since existing evaluation methods cannot be directly adopted to our setting, we also propose two novel evaluation protocols. The results on two large-scale re-id datasets (Market-1501, DukeMTMC-reID) demonstrate that our multi-camera method significantly outperforms baselines and other popular feature fusion schemes. Additionally, we conduct a comparative subject-based study of human operator performance. The superior operator performance enabled by our approach makes a compelling case for its integration into deployable video-surveillance systems. 
\end{abstract}

}

\maketitle

\IEEEdisplaynontitleabstractindextext

\IEEEpeerreviewmaketitle

\IEEEraisesectionheading{\section{Introduction}\label{sec:introduction}}

\IEEEPARstart{I}{n} recent times, the development of intelligent video surveillance platforms to monitor large crowded settings such as shopping malls, railway stations, airports etc. has become a priority to ensure public safety and security. A crucial component of such a platform is the \textit{person re-identification} (re-id) system. Given a query image, a re-id system searches through all the camera Field-of-Views (FoVs) and returns a per-camera ranked list of candidate matches. However, due to large variation in illumination, viewpoint, target resolution and other challenges arising from occluded targets, re-id methods are often unable to retrieve the correct match within a short enough ranked list. This imposes a significant burden on human operators of the surveillance system who now need to laboriously scan large lists per camera. The problem is further compounded when a large number of cameras are present. Such factors have kept person re-id an open problem in computer vision.

\begin{figure}[]
\centering
    \includegraphics[width=0.95\linewidth]{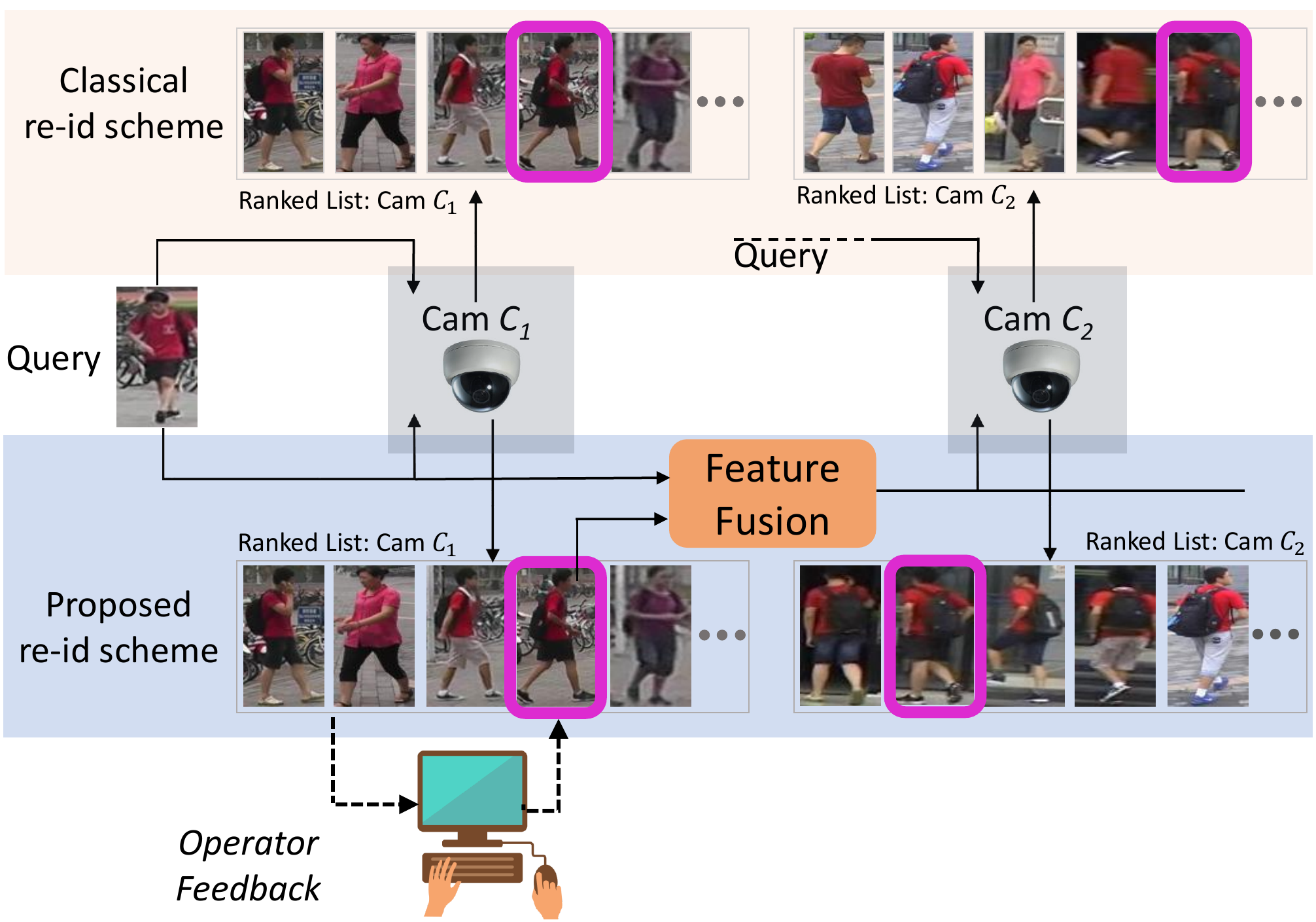}
    \caption{(Top) Classical re-id scheme where query image's feature representation is used to search each camera in the network independently. The retrieved lists are returned to the human operator. (Bottom) A real-world deployment scenario that motivates our proposed sequential re-id scheme where operator feedback regarding target sighting is utilized towards better re-id performance in an online fashion. In the figure, camera $C_1$ is queried first and ranked list of matches is obtained. The correct match (pink box) in the retrieved list is identified by operator and is subsequently fused with query image at feature level (orange block). This fused representation is used to query camera $C_2$. Notice that ranking of query target in $C_2$'s list is expected to improve in the sequential fusion-based approach unlike the classical version which cannot exploit operator inputs to improve subsequent queries.}
    \label{fig:pipeline}
\end{figure}

In a deployment scenario, it is fairly typical to observe a person in more than one camera FoV. Since each observation may provide complementary information, the human operator must seek the target in \textit{every} per-camera ranked list generated by a re-id system. If the target is identified in a particular list, the operator may choose to `label' the same via a simple haptic operation (e.g. touch or mouse-based selection). However, in a classical re-id scheme, the per-camera lists are generated \textit{independently}~\cite{koestinger2012large,liao2015person,ahmed2015improved} without taking actions of the human operator into account. In other words, target labeling by the operator in a subset of cameras cannot be leveraged to improve the ranking of the query target in the remaining set of cameras (see `Classical re-id scheme' in Figure \ref{fig:pipeline}). 

It is certainly desirable to exploit the complementary information on target appearance from multiple camera FoVs and consequent operator labeling. To this end, we propose a novel sequential and iterative approach which improves ranking of the target as additional cameras are queried across the network. Towards the success of our approach, we develop a sequential multi-camera fusion scheme. The fusion scheme operates on feature representations of candidate matches (see `Proposed re-id scheme' in Figure \ref{fig:pipeline}). Our approach has three major advantages. Firstly, it can accommodate an arbitrary number of cameras. Secondly, the fusion scheme is flexible enough to operate on cameras in any arbitrary order. Thirdly and crucially, our approach is designed to produce a \textit{monotonic} improvement in re-id performance as additional target labels from different cameras are fused. 

In addition, the proposed approach naturally aligns with the manner in which a human operator typically interacts with a re-id system. Therefore, it can be seamlessly integrated into deployable video-surveillance systems. The proposed approach is also designed as plug-and-play, i.e., it can be used atop any state-of-the-art camera pairwise feature estimation/metric learning method for re-id. Therefore, improvements in the camera-pairwise re-id approaches can be utilized and further extended within our framework. Concretely, we make the following contributions:
\begin{itemize}
    \item We propose a novel framework for utilizing feedback from human operators in a re-id pipeline deployed in a real-world scenario. In this proposed framework, observations from query target in a subset of cameras can be aggregated to obtain improved retrieval results for the remaining cameras in the network (Sec. \ref{sec:proposedmethod}).
    \item We propose a novel sequential feature fusion scheme and a training strategy that learns to achieve monotonic improvement in re-id performance as additional observations from the target are fused. (Sec. \ref{sec:fusrepimprovement}).
    \item To demonstrate the effectiveness of our approach, we define novel test protocols (Sec. \ref{sec:eval_protocol}) and perform extensive experiments (Sec. \ref{sec:results}) on two large-scale multi-camera benchmark datasets (Market-1501 \cite{zheng2015scalable}, DukeMTMC-reID \cite{ristani2016MTMC}).
    \item We perform comparative analysis of human operator performance obtained from interaction logs of a deployed re-id user interface to demonstrate the superiority and real-world feasibility of our approach.
\end{itemize}
We define two novel protocols to evaluate our proposed fusion framework. While both these protocols are directly motivated from the deployment scenario described in Fig. \ref{fig:pipeline}, they are also carefully modified to enable quantitative evaluation of fusion as well as comparison with traditional re-id approaches.

\section{Related Work}

The problem of person re-identification has been well studied over the last decade \cite{zheng-survey2016}. An important class of person re-id methods involve development of feature descriptions that are discriminative between different targets and exhibit robustness to variations in viewpoint, color, illumination etc. across different camera FoVs \cite{gray2008viewpoint,kviatkovsky2013color,martinel2012re,liu2012person,hu2013exploring,zhao2014learning,cheng2011custom}. 
Popular discriminative signature-based methods include ICT \cite{avraham2012learning}, SDALF \cite{bazzani2013symmetry}, saliency based methods \cite{zhao2013person,zhao2013unsupervised}
, hierarchical Gaussian descriptors \cite{matsukawa2016hierarchical} and many more. Besides these, a large volume of works have focused on camera-pairwise metric learning techniques~\cite{weinberger2009distance,hirzer2012relaxed,alipanahi2008distance,dikmen2010pedestrian}. 
Some widely used such techniques are LADF \cite{li2013learning}, RankSVM \cite{prosser2010person}, KISSME \cite{koestinger2012large}, LFDA  \cite{pedagadi2013local}, CFML \cite{alipanahi2008distance} and XQDA \cite{liao2015person}.

Recently, deep neural network based person re-id approaches have shown significant performance improvements by jointly learning the feature representation and the distance metric \cite{varior2016siamese,liu2017end,su2016deep,chen2017multi,li2017learning,wang2016joint,yi2014deep,li2014deepreid}. 
Unlike the classical hand-crafted techniques where the feature extraction and the metric learning methods were independently designed and cascaded, deep learning approaches jointly optimize for these two interconnected components, outperforming the \textit{non-deep} methods in the process. Many such methods solve re-id as a verification/binary classification problem. A popular approach involves Siamese networks with contrastive loss \cite{ahmed2015improved,varior2016gated}. In \cite{varior2016siamese}, LSTM modules were introduced into a Siamese network to model spatial dependencies between image parts. \cite{xiao2016learning} proposed a domain-guided dropout strategy to make the learned re-id model robust to inter-dataset variations. Even beyond Siamese, \cite{cheng2016person} provides an improved triplet loss for obtaining a more discriminative feature representation.
In datasets with large number of unique identities \cite{zheng2015scalable}, robust feature representations can be learned in an identification mode, i.e., training to map each image to an ID and using the learned feature embedding to associate unseen IDs during testing phase \cite{zheng-survey2016,zheng2016discriminatively,sun2017svdnet}. Specifically, in \cite{zheng-survey2016}, the authors implemented  a modified ResNet-50 \cite{he2016deep} model on Market-1501 \cite{zheng2015scalable} dataset under both identification and verification setup. We adopt the verification based protocol and baseline model in our experiments.

Recurrent Neural Networks have been used for feature aggregation in various video-based applications\cite{mclaughlin2016recurrent,xu2017jointly,yue2015beyond}.  Feature fusion for person re-id has also been considered, but in a multi-query set-up where multiple images of a target from the same camera are fused using simple pooling operations on feature representations \cite{zheng2015scalable}. Multi-camera fusion has been employed for object detection \cite{bhinge2017data}, tracking \cite{dockstader2001multiple} and activity classification \cite{hekmat2016multi}. While there are works in other fields with operator/human-in-the-loop frameworks, they essentially differ from our work in the manner in which the human feedback is made use of. For e.g, ~\cite{Cao2015} tries to learn similarity between face images from probe and gallery sets with human assistance. The work uses similarity labels as feedback from humans to iteratively embed the query into the learned feature set. Similarly, multi-camera feature fusion has been considered in the literature, but in a way unlike the proposed approach. Images from multiple cameras are used either in training or during inference to obtain a single fused representation which is then used for decision making (~\cite{harguess2009fusing,ristani2018features}). In contrast, the proposed fusion framework involves sequential fusion of inputs from multiple cameras. The current fused representation is used to query and retrieve images from gallery set and the retrieval features are then combined with the existing fused feature to obtain the subsequent query. To the best of our knowledge, ours is the first work to utilize operator feedback in such a sequential framework to perform fusion at the feature level. 

\section{Proposed Approach}
\label{sec:proposedmethod}
\begin{figure*}[!htbp]
\centering
    \includegraphics[width=0.63\linewidth,keepaspectratio]{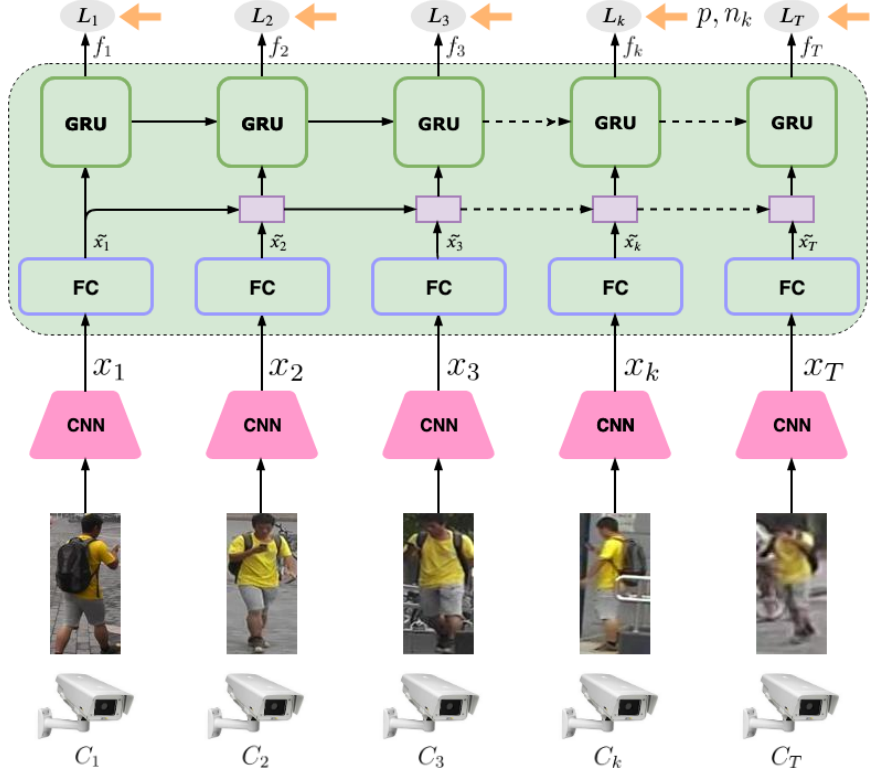}
    \caption{An illustration of our fusion architecture (Sec. \ref{sec:proposedmethod}). The baseline CNN features ($x_1,x_2,\ldots$) from camera images are fed to our fusion function. Purple boxes indicate mean-pooling of corresponding inputs. The fusion network is optimized via a novel loss formulation $L_k$, applied at each time step $k$, to improve the accumulated feature representation ($f_k)$. $p$ and $n_t$ are the representations for positive and negative instances. Note that for a given training sequence, the person id (anchor) and positive instance are held constant across the cameras, while negative instances vary. }
    \label{fig:GRU_Arch}
\end{figure*}

In this section, we lay out details of our method. We begin with a formal problem statement of our fusion approach (Section \ref{sec:problemstatement}). Having done so, we identify three key properties that need to be satisfied during the fusion process (Section \ref{sec:fusionfuncproperties}). We subsequently present the design of the fusion function (Sec. \ref{sec:grufusionfunc}) and our novel modifications to the default optimization procedure (Sec. \ref{sec:fusrepimprovement}), all designed to satisfy the properties mentioned previously. A GRU based implementation of the proposed fusion function is described in Sec.~\ref{sec:fusion_implementation_gru}.

\subsection{Problem Statement}
\label{sec:problemstatement}

Obtaining discriminative person-specific representations is a key component of any modern re-id approach. To obtain such representations, we follow the standard convention of fine-tuning pre-trained Convolutional Neural Networks (CNNs) on person re-id datasets for classification/verification task. For a given person image, we use the corresponding final, fully-connected layer's output of the fine-tuned CNN as the feature representation and employ $x$ or its subscripted variants to refer to the same.

Our problem can now be stated as follows: Suppose the total number of cameras is $T$ and the human operator has performed selection of the query target in $k \leqslant T$ cameras. Given the sequence of corresponding features $\{x_1, x_2, \ldots , x_k\}$, the aim is to learn a fusion function $\mathcal{F}$ that integrates operator feedback and produces an optimal fused representation $f_k$, i.e. $f_k = \mathcal{F}(x_1, x_2, \ldots ,x_k)$\footnote{Please note the distinction between fixed feature representations ($x$) obtained from CNN and the `learnt' fused representations ($f$) produced by our fusion function.}. 

\subsection{Desired Properties of the Fusion Function}
\label{sec:fusionfuncproperties}

The number of camera FoVs in which a query is visible can vary from target to target. Therefore, the fusion function $\mathcal{F}$ must be capable of handling a variable number of input feature representations. In addition, images of the same target in different camera FoVs often provide complementary visual information. Hence, a proper fusion of these image features should produce a more robust and holistic feature representation that leads to a better re-id accuracy/mAP. To achieve these aims, the proposed fusion approach must ideally satisfy the following properties: 

\begin{enumerate}
    \item $\mathcal{F}$ must be able to process camera (feature) sequences of variable lengths, i.e. $k$ can vary from target to target. 
    \item As the number ($k$) of feature representations being aggregated increases, the fused representation $f_k$ should improve, i.e. enable sustenance or increase in re-id accuracy.
    \item $\mathcal{F}$ should be invariant to relative ordering in the input feature sequence, i.e. the order in which cameras are considered should not matter.
\end{enumerate}

\subsection{Design of the Fusion Function}
\label{sec:grufusionfunc}

A feature fusion module can be designed in a number of possible ways. Among the popular early fusion/feature fusion techniques, mean and max pooling (element-wise for the feature vectors) can be suitable candidates for our fusion function as both of these satisfy the desired properties-1 and 3 by design. However, these methods do not necessarily guarantee the property-2, i.e., the fused representations resulting in sustenance or improvement in re-id accuracy when longer sequences of features are input to the fusion function. Towards this, the function should be designed such that it contains learnable parameters and the desired properties (e.g., property-2) can be implicitly enforced via minimization of a suitable cost over these parameters. Most recurrent models (e.g. recurrent neural nets) would be classified under this category of functions. In the current and the following subsection, we describe the design of the suitable cost functions for optimal estimation of the fusion function parameters.

During the training phase, we require $\mathcal{F}$ to transform the sequence of image features $\{x_1, x_2, \ldots ,x_k\}$ from the $k$ different cameras to a corresponding sequence of fused representations 
$\{f_1, f_2, \ldots , f_k\}$ (Sec. \ref{sec:problemstatement}). To achieve this, the image features are first transformed to an embedding of pre-defined dimension to obtain $\{\Tilde{x}_1, \Tilde{x}_2, \ldots \}$. To increase the robustness of the fusion process, $\{\Tilde{x}_1, \Tilde{x}_2, \ldots \}$ up to and including current camera index $t$ are mean-pooled (purple boxes in Fig.~\ref{fig:GRU_Arch}) and fed as input to a recurrent function block. 

Suppose we choose an image from a training sequence and define it as the \textit{anchor}. We define \textit{positive} instances as those training images having the same id as that of the anchor and \textit{negative} instances as those images whose id differs from anchor's id. Ideally, we require that a positive instance's feature representation be closer to anchor's representation than the negative's representation. 

This objective can be achieved via minimization of a hinge-style \textit{triplet} loss~\cite{ding2015deep,cheng2016person,schroff2015facenet,HermansBeyer2017Arxiv} defined on the anchor, positive and negative instance representations.

\begin{equation}
    L^{(tri)} = \sum_{\{f, p, n\}}\text{max}(0, \|f - p\|_2 - \|f - n\|_2 + m)
\label{eqn:loss_gru_1}
\end{equation}

where $m$ is the margin. 

In our setting, we set up the triplet loss $L^{(tri)}_t$ for each camera index $t$ wherein
the fused representation $f_t$ serves as the anchor. The choice of positive instances is limited, being confined to same sequence or at the most a handful of other sequences. We omit the camera corresponding to the positive instance during the fusion process since images from the same camera have high similarity. We also choose to keep the positive instance fixed for \textit{all} indices of a given training sequence. The negative instance for each index is chosen using hard mining within a given training mini-batch~\cite{HermansBeyer2017Arxiv}. To enable comparison with the fused feature, the positive and negative instances are processed by the fusion module $\mathcal{F}$ for a single time-step to obtain the corresponding features $p$ and $n_t$. The triplet loss at index $t$ is defined as:
\begin{equation}
    L^{(tri)}_t = \sum_{\{f_t, p, n_t\}}\text{max}(0, \|f_t - p\|_2 - \|f_t - n_t\|_2 + m)
\label{eqn:loss_gru}
\end{equation}

We use soft-margin formulation as an approximation to the hinge loss~\cite{HermansBeyer2017Arxiv} as follows:
\begin{equation}
    L^{(tri)}_t = \sum_{\{f_t, p, n_t\}} \text{ln}(1 + e^{\|f_t - p\|_2 - \|f_t - n_t\|_2})
\end{equation}

\subsection{Monotonic Representation Improvement}
\label{sec:fusrepimprovement}

To specifically address the requirement of progressive improvement in the quality of the fused representation $f_t$ (property-2 in Section \ref{sec:fusionfuncproperties}), we introduce an additional per-index loss term called \textit{monotonicity loss} (m-loss). m-loss is formulated as a sum of zero-margin hinge losses as follows:
\begin{equation}
\label{eqn:loss_rank}
\resizebox{.99\hsize}{!}{
    $L^{(mon)}_t = \sum_{\{f_t, p, n_t\}}\text{max}\big(0,\ d(f_t,p)-d_p^{*(t)}\big) + \text{max}\big(0,\  d_n^{*(t)}-d(f_t,n_t)\big)$ 
    }
\end{equation}
where $d(.)$ is the euclidean distance metric. $d_p^{*(t)}$ and $d_n^{*(t)}$ are defined as follows:
\begin{subequations}
\label{eqn:m_loss}
    \begin{align}
        d_p^{*(t)} &= \min_{\tau\in\{1,2,\dots,t-1\}} d(f_\tau,p)\\
        d_n^{*(t)} &= \max_{\tau\in\{1,2,\dots,t-1\}} d(f_\tau,n_\tau)
    \end{align}
\end{subequations}

Eq.~\ref{eqn:loss_rank} and eq.~\ref{eqn:m_loss}(a) ensure that the fused representation at step $t$ is closer to the positive instance than all the fused representations till index $t-1$. Also, the negative instance is chosen using hard-mining within a mini-batch and $d_n^{*(t)}$ is chosen as the maximum of distances from the fused representations to the corresponding negative samples. Therefore eq.~\ref{eqn:loss_rank} and eq.~\ref{eqn:m_loss}(b) enforce $f_t$ to be farther from all negative samples in the mini-batch compared to any fused representation till step $t-1$.

The total loss at each time step $t$ is formulated as a weighted combination of the triplet loss and the monotonicity loss, i.e. 
\begin{equation}
L_t = L^{(tri)}_t +  \lambda \lambda^{R}_{t} L^{(mon)}_t
\label{eqn:totalloss}
\end{equation}

While $\lambda$ is fixed for all indices $t$, $\lambda^{R}_{t}$ is obtained using a linear weighting scheme to give more importance to monotonicity loss for longer sequences. For a sequence of length $T$, $\lambda^{R}_t = t / T$. Overall, the proposed loss formulation is designed to ensure a decoupled optimization of the two desired properties -- low triplet loss when a new feature representation is aggregated and monotonic improvement in fused feature representation.


\subsection{Implementation of the Fusion Module}
\label{sec:fusion_implementation_gru}

To meet the requirements for the fusion function as described above, we judiciously design $\mathcal{F}$ around as a recurrent neural network. Specifically, out of many choices (RNNs, LSTMs, GRUs etc.) for the recurrent architectures, we choose to use a Gated Recurrent Unit (GRU) (Fig.~\ref{fig:GRU_Arch})~\cite{Chung:2015:GFR:3045118.3045338} -  a popular Recurrent Neural Network architecture (Sec. \ref{sec:grufusionfunc}). In GRU, the following set of transformations are applied at each index $t$ of the sequence:

\begin{subequations}
\label{eqn:gru_eq}
\begin{align}
    r_t   & = \sigma(W_{rx}x_t + W_{rh}h_{t-1} + b_r)\\
    z_t   & = \sigma(W_{zx}x_t + W_{zh}h_{t-1} + b_z)\\
    s_t & = \text{tanh}(W_{hx}x_t + W_{hh}(h_{t-1}\odot r_t) + b_h)\\
    h_t & = (1-z_t)\odot h_{t-1} + z_t\odot s_{t}
\end{align}
\end{subequations}

Here, $\odot$ represents element-wise multiplication and $\sigma$ represents the sigmoid function. $h_t$ is 
formulated to serve as an effective feature representation for the input feature sequence $\{x_1, x_2, \ldots , x_{t}\}$ seen until that point, i.e., $f_t$ = $h_t$. The intermediate transformations $r_t,z_t,s_t$ are formulated such that the GRU effectively fuses only helpful aspects of the input and ignores the rest. Our design choice of GRU is significantly motivated by this property. Note that the subscripted $W$'s and $b$'s are shared across all the sequence indices and form the trainable parameters of the GRU. 

\subsection{Training and Testing}
\label{sec:trainandtest}

The sequence-loss for GRU is computed as an average across per-index total loss (eq. \ref{eqn:totalloss}). During the fusion network training, we nominally fix an input camera sequence ordering and the inputs to the GRU are obtained on the basis of this ordering. We emphasize that the choice of ordering is arbitrary. In fact, we shall show later that the camera ordering has negligible effect on re-id performance (Sec. \ref{sec:results}). This result also implies that the fusion function satisfies the third property from the desirable properties of an ideal fusion function (Sec \ref{sec:fusionfuncproperties}).

In the testing phase, query images from multiple cameras are considered for fusion. We use the hidden state $h_k$ of the GRU at the last camera index (Eq.~\ref{eqn:gru_eq}(d)) as the fused feature $f_k$. Since the ids of images in the gallery set are unknown, it is not possible to obtain a fused representation for them. To enable comparison between query and gallery features, we construct a sequence by repeating the gallery image and use it as the input to the GRU. Additional details on this procedure are presented in Sec.~\ref{sec:eval_protocol}.

\vspace{1mm}
\noindent \textbf{Other Fusion Functions:} We explore \textit{mean-pooling} and \textit{max-pooling} of features as two alternative fusion functions. As discussed earlier in this section, both these functions (with no trainable parameters) satisfy the desired properties 1 and 3 by design. These pooling operations are performed in ways similar to multi query setting for person re-identification~\cite{zheng2015scalable} to obtain the fused representations. We present a detailed comparative evaluation of the fusion functions in Sec. \ref{sec:Experiments}. We also show how the early/feature fusion based sequential re-id compares in performance with two late-fusion approaches (Sec.~\ref{sec:latefusion}).

\section{Experiments}
\label{sec:Experiments}
\subsection{Datasets}
 
Since the focus of the work is on fusion of features from multiple cameras, we evaluate performance on datasets with a minimum of three cameras in the network. We report our results on two such datasets, Market-1501 and DukeMTMC-ReID, which contain $6$ and $8$ cameras respectively. 

\begin{figure}
        \begin{subfigure}[b]{0.48\linewidth}
                \includegraphics[width=\linewidth]{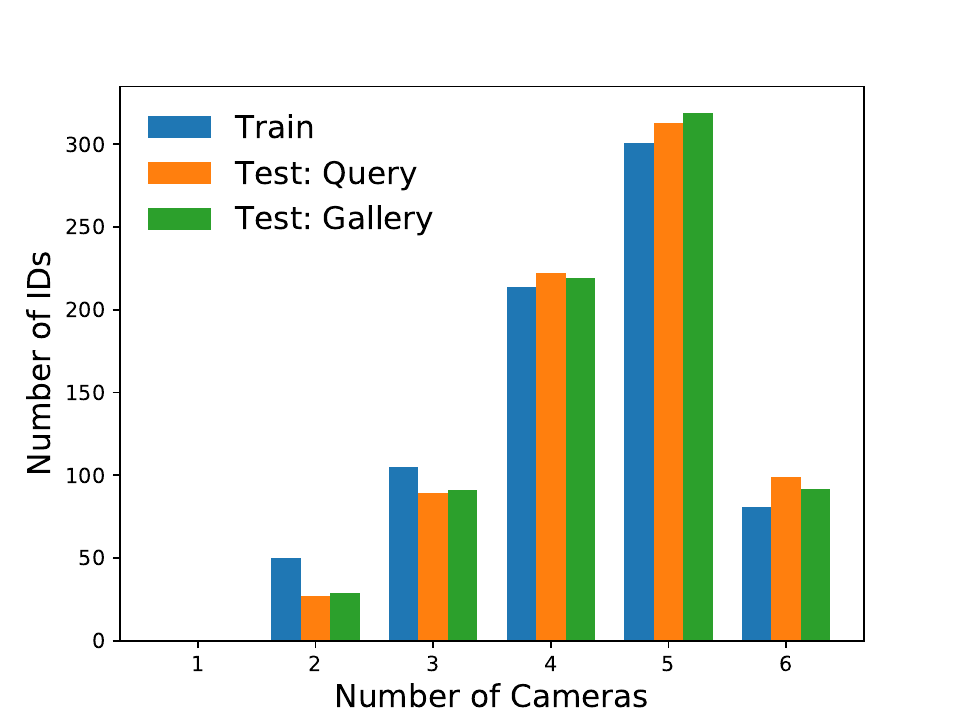}
                \label{fig:histmarket}
        \end{subfigure}
        \begin{subfigure}[b]{0.48\linewidth}
                \includegraphics[width=\linewidth]{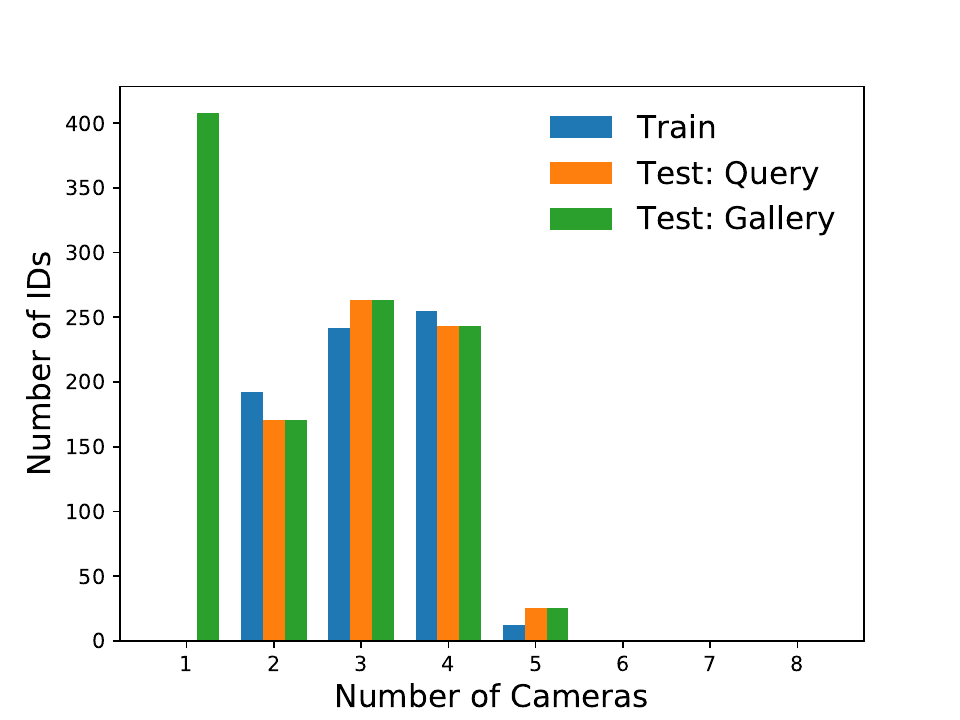}
                \label{fig:histduke}
        \end{subfigure}
        \caption{Histogram of maximum number of cameras each target is observed in Market-1501~\cite{zheng2015scalable} (left) and DukeMTMC-reID~\cite{Zheng_2017_ICCV} (right) datasets. In DukeMTMC-reID, there are very few samples with query sequence lengths greater than $4$.}
        \label{fig:histogram}
\end{figure}

\vspace{1mm}
\noindent \textbf{Market-1501}~\cite{zheng2015scalable}: This dataset has $12{,}936$ images from $751$ IDs in the training set and another $750$ test IDs with $3{,}368$ and $19{,}732$ images in the query and gallery sets respectively. Each ID is present in a minimum of two and a maximum of six cameras (see left plot in Fig.~\ref{fig:histogram}). The gallery set has multiple instances of an ID from a camera while the query set has only one. All the images are of dimensions $128 \times 64$.

\vspace{1mm}
\noindent \textbf{DukeMTMC-ReID}~\cite{Zheng_2017_ICCV}: This dataset is organized similar to Market-1501. It has $702$ IDs each in the train and test sets. There are $16{,}522$, $2{,}228$ and $17{,}661$ images in train, query and gallery sets respectively. All the images are obtained using manually annotated bounding boxes. In the training set, each ID is present in a minimum of $2$ and a maximum of $6$ cameras, even though the network has $8$ cameras (Fig.~\ref{fig:histogram}). The gallery set has 408 distractor IDs, not present in more than one camera FoV.

\subsection{Implementation Details}

\subsubsection{Feature Extraction}
For our experiments, we use ResNet-50~\cite{he2016deep} and AlexNet~\cite{krizhevsky2012imagenet} as the base (per camera image) CNN feature extractor models. Note that these choices are nominal and any off-the-shelf model can be used as the baseline feature extractor. 

For the ResNet-50 baseline, we use the network pre-trained on ImageNet~\cite{deng2009imagenet} for fine-tuning on reID datasets. An additional fully-connected (FC) layer is used at the end of Pool-5 layer of ResNet-50 to reduce the feature dimension to $512$. For the AlexNet baseline, we remove Local Response Normalization and employ batch-normalization at every layer before the non-linearity. Similar to the ResNet-50 set-up, the output embedding dimension is set to $512$. During the baseline network training, dropout with rate $0.5$ is employed for the fully-connected layers. We use Adam optimizer with an initial learning rate of $0.0001$. $\beta_1$ and $\beta_2$ parameters in the optimizer are set to $0.9$ and $0.999$ in all experiments. As done in~\cite{HermansBeyer2017Arxiv}, the learning rate is decreased as the training progresses according to the following schedule:

\begin{equation}
\label{eqn:sched}
        \epsilon(t) = \begin{cases}
                          \epsilon_0 &\text{if} \ t \leqslant t_0\\
                          \epsilon_0 \times 0.001^{\bigg(\dfrac{t-t_0}{t_1-t_0}\bigg)} &\text{if} \ t_0 \leqslant t \leqslant t_1
                      \end{cases}
\end{equation}

Here, $\epsilon_0$, $t_0$ and $t_1$ are set to $0.0001$, $15000$ and $25000$ respectively. 

The input dimensions for ResNet-50 and AlexNet are fixed to $256 \times 128$ and $227 \times 227$ respectively and the input images are accordingly resized. To maintain the aspect ratio of input in ResNet-50, the pooling layer is modified to enable an input of dimension $256 \times 128$. Following~\cite{krizhevsky2012imagenet}, we augment 
our training set with $5$ random crops and their mirrored images. The size of crop is set to $89$\% of the original image size.

\subsubsection{Fusion Function}
The GRU is initialized with random weights and hidden state length is set to  $512$ in all our experiments. As in CNN training, we use the Adam optimizer to perform gradient descent. 
For the experiments with monotonicity loss (Sec. \ref{sec:fusrepimprovement}), the weighting factor $\lambda$ is calculated using the scheduling scheme similar to that in Eq.~\ref{eqn:sched} ($\epsilon,\epsilon_0$ replaced with $\lambda,\lambda_0$) with $\lambda_0$ equal to $0.01$.

\subsection{Evaluation Protocols}
\label{sec:eval_protocol}
In the protocol generally followed for evaluation in multi-camera setting~\cite{zheng-survey2016}, single query and single gallery sets are used irrespective of number of cameras in the network. The images from all the cameras are binned together in the gallery and for a given query, predictions from the same camera are treated as inadmissible, i.e. not considered for evaluation. In our work, we tackle the novel task of cross-camera fusion which requires a minimum of two camera inputs into the fusion function and at least one gallery camera to compare the fused representations against. This setting is different from traditional protocols and hence the existing evaluation procedures cannot be directly adopted. 

Therefore, we modify the traditional protocols under the constraints present in re-id datasets and propose two new evaluation protocols -- Variable Set Protocol (VSP) and Fixed Set Protocol (FSP). These two protocols are explained in detail in the following sections. The suitability aspect towards evaluation of our proposed framework and design justifications for each of these protocols are also discussed in detail.

\begin{figure}
    \centering
    \includegraphics[width=\linewidth]{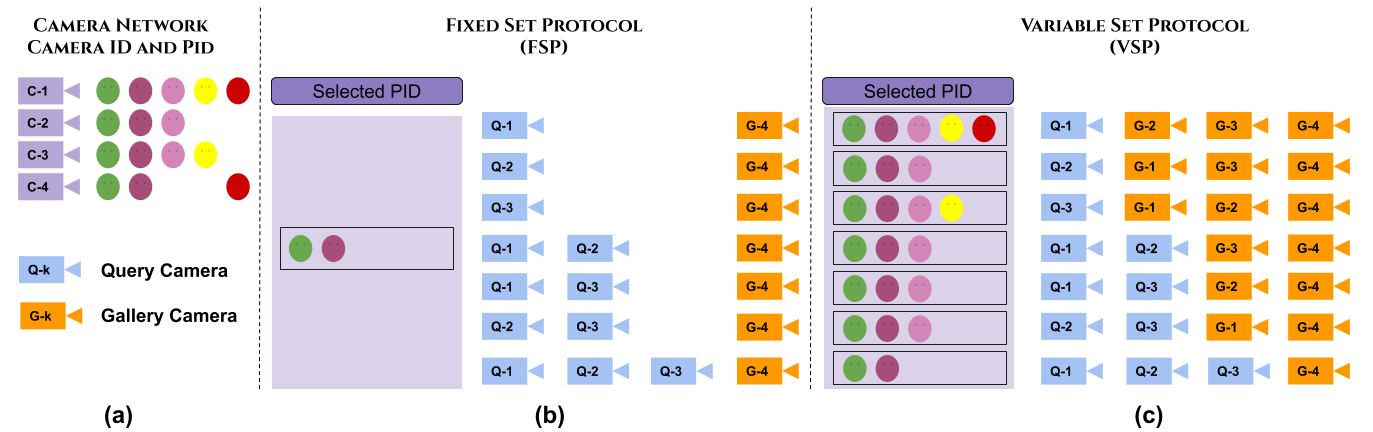}
    \caption{Comparison of camera subset and ID selection in FSP and VSP evaluation protocols. An illustration of the protocols is provided for a sample camera network with four cameras and five unique identities, denoted by different colored circles. Different query-gallery camera subsets and corresponding IDs for querying are shown for each protocol. Note that for VSP, only a subset of all possible camera combinations are shown for brevity.} 
    \label{fig:eval_prot_compare}
\end{figure}

\subsubsection{Variable Set Protocol (VSP):}
\label{sec:vsp}

Note that we require a comparison of the proposed approach with traditional re-id methods used as baselines in this work (along with alternative fusion approaches). Therefore, we develop a protocol characteristically very similar to the traditional re-id evaluation setups, while suitably modified to align with our sequential re-id philosophy. For this, we partition the dataset into two sets of observations, which are not only mutually exclusive in terms of their image contents, but are also disjoint in terms of camera field-of-views (FoVs) from which the observations are sourced.

\begin{figure}[!t]
    \centering
    \begin{minipage}{\linewidth}
        \begin{minipage}{0.49\textwidth}
            \centering
            \includegraphics[scale=0.28]{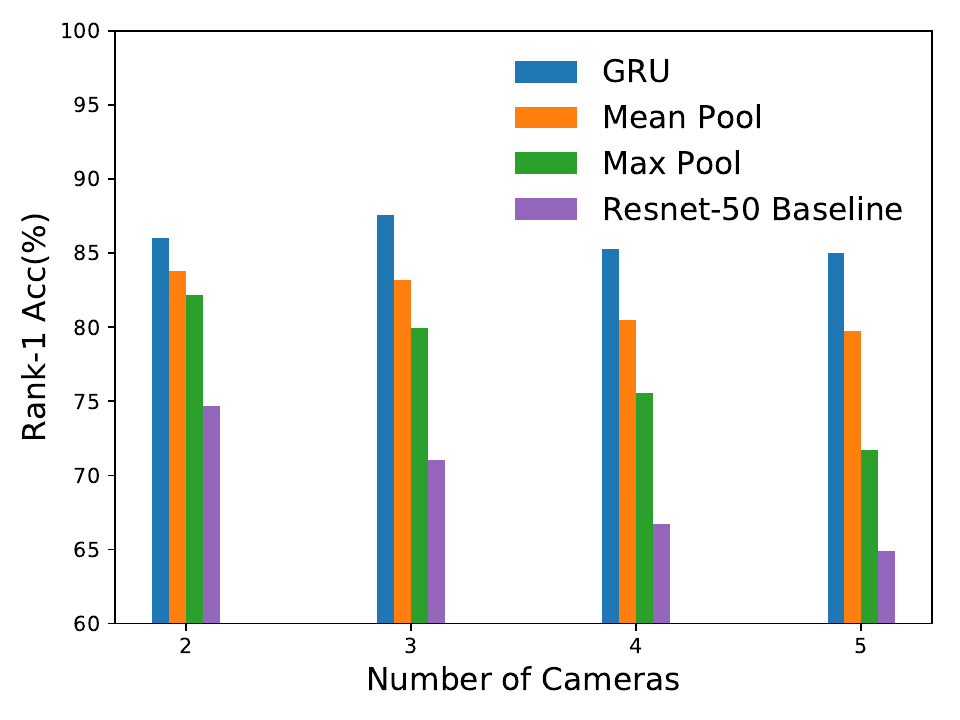}
        \end{minipage}
        \begin{minipage}{0.49\textwidth}
            \centering
            \includegraphics[scale=0.28]{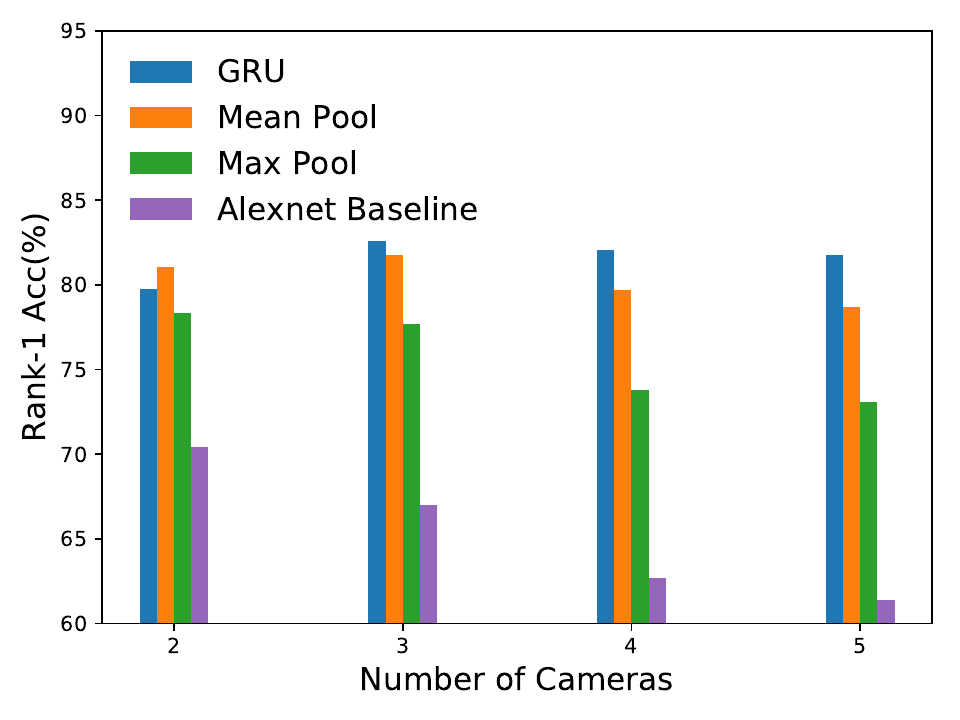}
        \end{minipage}
    \end{minipage}
    \caption{Rank-1 accuracies for Variable Set Protocol on Market-1501 dataset with ResNet-50 (left) and AlexNet (right) CNN baselines. GRU based 
            fusion outperforms mean and max-pooling based fusion methods. All fusion techniques are significantly better than the CNN baselines.}
    \label{fig:prot2_M}
\end{figure}

Let $C$ be the set of cameras present in the network (Fig.~\ref{fig:eval_prot_compare} a). A subset of $C$ is considered as the gallery camera set $\{{G}_{C}\}$. The complementary set of $\{{G}_{C}\}$ is considered to be the query camera set $\{{Q}_{C}\}$. For evaluation, a set of query person IDs from $\{{Q}_{C}\}$ are selected such that they are present in a minimum of one camera in $\{{G}_{C}\}$ (Fig.~\ref{fig:eval_prot_compare} c). This procedure is repeated for all possible gallery camera sets. Finally results are averaged over query subsets of same cardinality. The total number of such query-gallery combinations is given by $N = \sum_{i=1}^{n-1} \binom{n}{i} = 2^{n}-2$ where $n$ is the number of cameras in the network.  
Note that the size and contents of both the query and gallery sets change based on the selection of number of cameras for fusion (each cluster of bars in Fig.~\ref{fig:prot2_M}). Hence, it is not possible to compare the performance of fusion function for different lengths of sequences across these different sets of partitions, following this protocol. We have designed the FSP protocol specifically to study this.

\subsubsection{Fixed Set Protocol (FSP):}
\label{sec:fsp}
We design this protocol to specifically evaluate the fusion approaches. In an example deployment setup, shown in Fig.~\ref{fig:pipeline},
the retrieved image from a camera is fused with the query and a \textit{different} camera with non-overlapping field-of-view is queried to obtain the subsequent retrieval. This procedure is repeated to obtain fusions of growing number of images until all the cameras in the network are exhausted. Thus the query set before and after a certain fusion step changes in size and contents. To evaluate the proposed fusion scheme, one needs to freeze the gallery set to allow for a fair comparison of fusion performance across different query subsets. This practical constraint of having the same evaluation setting for comparing different lengths of fusion compels us to design an alternative evaluation protocol, termed `Fixed Set Protocol' (FSP).  
Instead of fusing query images with the retrieved images from gallery, images of the same target from the query camera subsets are used for fusion and then evaluation of the performance of this fusion is performed on the fixed gallery set ($\{G_{C}\}$).
Note that the inputs to be fused are selected using ground truth labels, 
and can be thought of as the \lq{}virtual\rq{} human operator's feedback.
To evaluate the fusion function's performance, we consider all possible subsets of 
cameras within the query set ($\{Q_{C}\}$), starting with one camera (no fusion) and progressively 
increasing until $N_Q$, the number of cameras within the query set. For further clarifying these motivations as well as towards better understanding of this protocol, refer to Fig. ~\ref{fig:eval_prot_compare} (a,b).
Let there be four cameras in a network, numbered $1$, $2$, $3$, $4$. Cameras $1$, $2$ and $3$ are query cameras, fused in this particular order and the camera $4$ constitutes the gallery. In the first step, query from camera $1$ is combined with a \lq{}retrieved\rq{} (we use ground truth labels to simulate retrieval by human operator) image from camera $2$ to query the gallery set. Subsequently input from camera $3$ is combined with the previous fused representation to again query the same gallery set. This procedure is repeated for all possible query camera combinations. The total number of such possible query camera combinations in any camera network is $N = |\mathcal{P}(Q_C)|-1$ where $\mathcal{P}(S)$ and $|S|$ are the 
power set and the cardinality of $\{S\}$ respectively. Note that we choose only those IDs which are present in all the cameras 
in both $\{Q_{C}\}$ and $\{G_{C}\}$ so as to enable fusion in any query camera subset (Fig.~\ref{fig:eval_prot_compare}). Thus the set of query IDs is \textit{fixed} for a given gallery set regardless of the query subset used, rendering the metrics for different query subset combinations comparable.

In the test phase for both protocols, feature fusion is performed only on the query subset of the dataset. To enable comparison of query and gallery features during testing, we mimic the multi-camera scenario by constructing a sequence of repeated gallery image features. Our decision is motivated by the fact that our fusion function is optimized for sequences and also by better performance observed in practice. We empirically set the number of gallery image repetitions to be same as the query sequence length. In the following sub-sections, we report results for the GRU based fusion function trained only with triplet loss (termed GRU) in both VSP and FSP protocols. To show the efficacy of m-loss, we would need to compare performance across different query sequence lengths and thus report results for GRU trained with both triplet and m-loss (termed GRU+m-loss) only on the FSP protocol. 

Overall, the proposed VSP and FSP protocols enable us to evaluate a realistic deployment scenario and quantitatively compare such a scenario with traditional baseline schemes. More specifically, FSP has been designed to compare the utility of fusion and the proposed `m-loss' (Section \ref{sec:fusrepimprovement}) across variable-length observation sequences. In contrast, VSP is aimed broadly towards comparison of the proposed GRU based fusion framework with traditional re-id models used as baselines in this work. 

\subsection{Results}
\label{sec:results}

\subsubsection{Baseline CNN performance} 

Table~\ref{tab:baseline} shows the rank-1 and mAP metrics for the ResNet-50 and AlexNet CNN baselines on Market-1501 and DukeMTMC-ReID datasets. The ResNet based network significantly outperforms the AlexNet based network. The above pre-trained baseline networks are used as the feature extractors for the fusion module in all our experiments. Since ResNet based network achieves better retrieval performance, we primarily show results using the ResNet baseline.  
\begin{table}[t]
    \centering
    \begin{tabular}{|l|c|c|c|c|}
    \hline
    \multirow{2}{*}{Architecture} & \multicolumn{2}{c|}{Market-1501}                        & \multicolumn{2}{c|}{DukeMTMC-reID}   \\ \cline{2-5}
                                  & Rank-1 $\uparrow$        & mAP  $\uparrow$      & Rank-1  $\uparrow$       & mAP $\uparrow$    \\ 
    \hline
    ResNet-50                    & 73.63 & 48.74   & 60.86  & 39.79  \\ 
    AlexNet                      & 67.1 & 44.34   & 56.87 & 34.21 \\
    \hline
    \end{tabular}
    \caption{Classification-based Baseline CNN performance.}
    \label{tab:baseline}
\end{table}

\begin{table}[!t]
\resizebox{0.49\textwidth}{!}
{%
    \centering
        \begin{tabular}{|l|c|c|c|c|c|c|c|}
            \hline
            \multirow{2}{*}{Architecture}     & \multicolumn{3}{c|}{mAP for FSP}              & \multicolumn{3}{c|}{mAP for VSP}                      \\ \cline{2-7}
                                              & {Proposed}  & {Mean-pool}  & {Max-pool}   & {Proposed}    & {Mean-pool}       & {Max-pool}      \\   
            \hline                                                       
            ResNet-50+Fusion                         & \textbf{75.87}        & 67.88   & 64.52       & \textbf{71.82}    & 64.65             & 61.29   \\ 
            AlexNet+Fusion                           & \textbf{67.79}        & 67.13   & 63.65       & \textbf{63.71}    & 63.10             & 59.44   \\
            \hline                                                       
            \end{tabular}                                                
}
\caption{Comparison of averaged mAP on Market-1501 with unit length galleries. The proposed fusion methodology consistently outperforms the other feature fusion techniques. Note that the baseline CNN performance for VSP is $54.31\%$ and $51.05\%$ respectively for ResNet-50 and Alexnet based architectures.}
    \label{tab:comp_map}
\end{table}

\begin{figure}[t]
    \begin{center}
        \includegraphics[width=\linewidth]{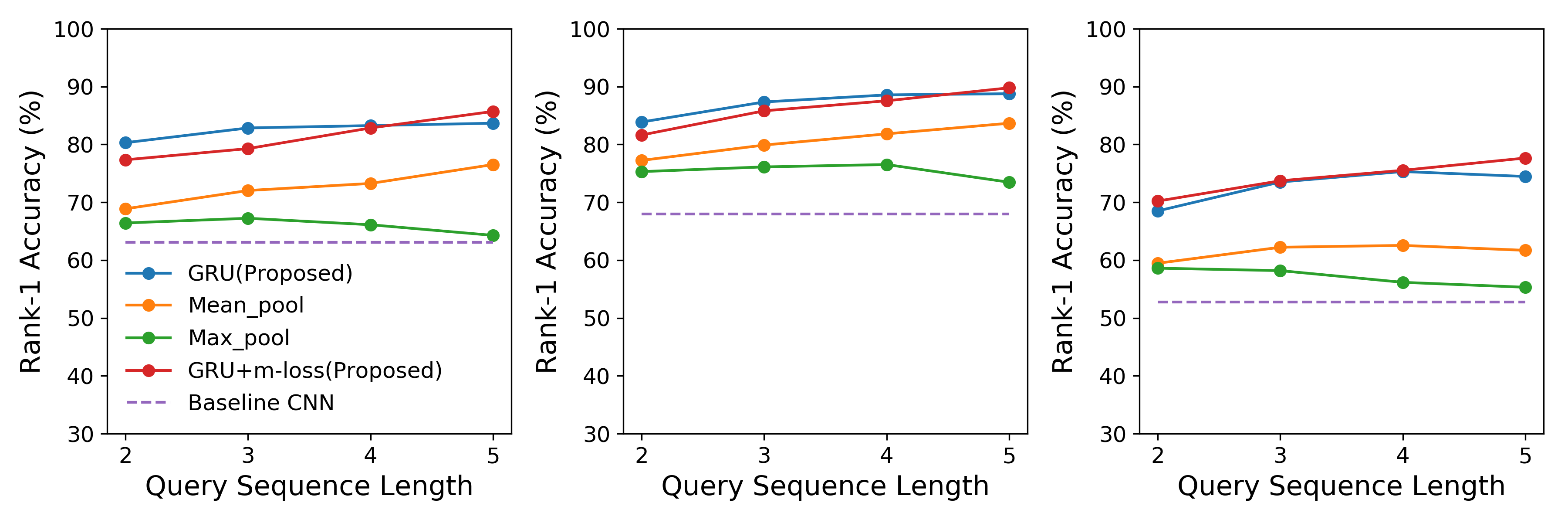}
    \caption{Effect of query sequence lengths on accuracy. Rank-1 accuracies on three different gallery sets are shown for fusion with ResNet-50 baseline on Market-1501 dataset. Feature fusion using GRU performs significantly better than other fusion techniques in rank-1 accuracies, while m-loss helps in maintaining monotonicity with increasing sequence lengths}
    \label{fig:prot1_all}
\end{center}
\end{figure}

\subsubsection{Results with VSP} 
\label{subsec:vsp_results}
The results for VSP on Market-1501 are shown in Fig.~\ref{fig:prot2_M}. Since the baseline feature extractor methods take in inputs from only one camera at a time, we independently query from each of the cameras in the query set. The scores are computed for each of these individual queries and their average is considered for comparison with feature fusion based methods. In this protocol, we report results for GRU based fusion function trained with just the triplet loss. 

For better representation, we average the results based on the number of cameras present in the query set. From the results (Fig.~\ref{fig:prot2_M}), we observe that our approach (fusion of queries) performs significantly better than baseline --  for ResNet-50, on average, fusion outperforms baseline by $\mathbf{13.5\%}$ and mean-pool based fusion by $\mathbf{3.6\%}$ in Rank-1 accuracy. The mAP performances (Table~\ref{tab:comp_map}) are more noteworthy with $\mathbf{17.5\%}$ and $\mathbf{7.2\%}$ improvement over baseline CNN and mean-pool based fusion respectively. In the case of AlexNet as baseline CNN, mean-pool based fusion performs slightly better than our approach for sequences of length two. However, as the number of cameras increase, our approach outperforms all other approaches, thereby satisfying a design objective of our fusion function. Additionally, the figures show that the improvement obtained using feature fusion increases as more query cameras are considered, as expected.

\begin{table}[!t]
    \centering
    \begin{tabular}{|l|c|c|c|c|}
    \hline
    \multirow{2}{*}{Architecture} & \multicolumn{4}{|c|}{Rank-1 @ Query Sequence Length}\\ \cline{2-5} 
                                  & {2}     & {3}      & {4}       & {5}           \\ 
    \hline
    ResNet-50+Fusion                    & 79.37        & 83.58           & 84.29              & 85.01  \\ 
    AlexNet+Fusion                      & 71.21        & 76.88           & 80.14              & 81.77  \\
    \hline
    \end{tabular}
      \caption{Comparison of averaged FSP rank-1 accuracies of ResNet-50 and AlexNet based fusion with varying query set lengths. Evaluation is performed on the GRU+m-loss based fusion. Results are averaged over all six unit length gallery camera sets.}
    \label{tab:prot1_resnet_alexnet_compare}
\end{table}

\subsubsection{Results with FSP} 
\label{subsec:fsp_results}
\begin{figure}
    \centering
    \includegraphics[width=0.5\linewidth]{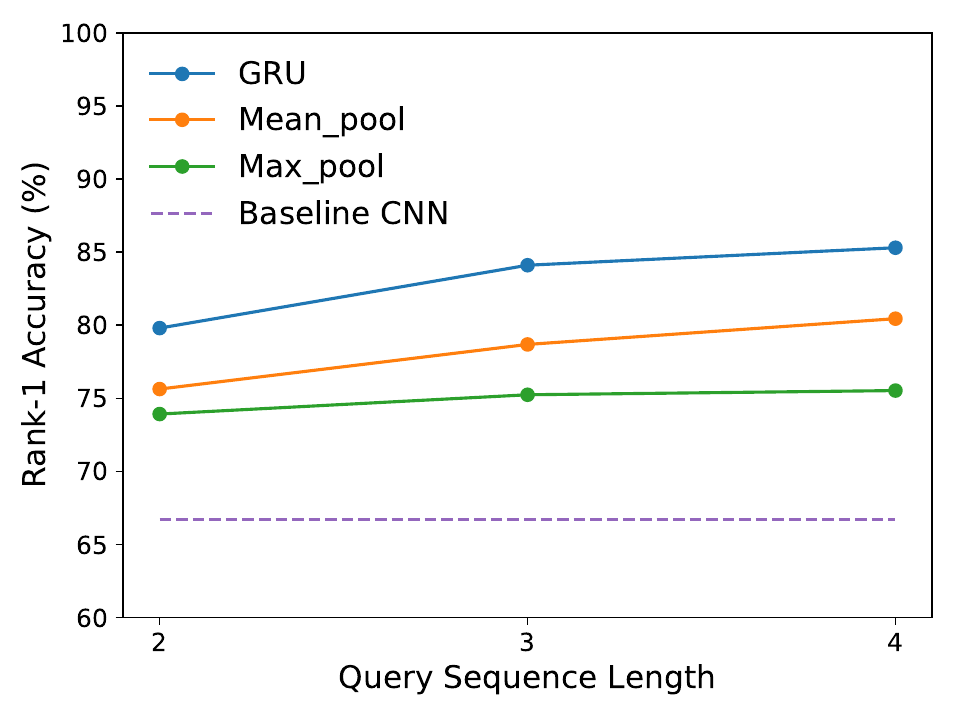}
    \caption{Averaged FSP results for size 2 gallery sets using ResNet-50 based network on Market-1501~\cite{zheng2015scalable} dataset. The fusion based approaches result in monotonic improvement with increase in query sequence length. The proposed GRU based fusion significantly outperforms the other fusion techniques.}
    \label{fig:gal_vary_market}
\end{figure}

To show the efficacy of fusion, we compare fusion performance for varying query sequence lengths with fixed gallery sets. The query sequence length refers to the cardinality of the query camera combination. In this protocol, we compare both the GRU (triplet only) as well as the same trained using the additional m-loss to show the utility of the latter as the length of sequence of observations to be fused increases. Table~\ref{tab:prot1_resnet_alexnet_compare} presents the comparison of the proposed fusion based re-id using ResNet-50 and AlexNet baselines on Market-1501 dataset using FSP for different query sequence lengths. The gallery camera set length is fixed to one. Hence, at most five images can be used for feature fusion. The average rank-1 accuracies over six such galleries is shown in the table. ResNet-50 based fusion network performs significantly better due to better baseline features. Hence, in the remaining experiments on FSP, we present results mainly on ResNet-50 architecture. 
The effect of number of query images on fusion accuracy can also be viewed in Fig.~\ref{fig:prot1_all}. The monotonic trend of accuracies with increase in number of query cameras holds in the case of GRU alone, but is further enhanced when trained with m-loss, leading to improved accuracy at the later time-steps. On an average, our fusion approach achieves $\mathbf{5.8\%}$ improvement in Rank-1 accuracy over mean-pooling. Table~\ref{tab:comp_map} provides a comparison of mAP with ResNet-50 and AlexNet architectures on Market-1501. For ResNet-50, our fusion approach outperforms mean-pool based fusion in mAP by about $\mathbf{8\%}$. The significant improvement in mAP indicates that the fused representation is able to effectively combine images, leading to better low-rank retrievals. The results also crucially highlight the advantage of our GRU-based fusion over simple pooling approaches. 
Fig.~\ref{fig:gal_vary_market} presents averaged FSP results on gallery sets with two cameras on Market-1501 dataset. Since there are two cameras in the gallery set, the maximum possible query sequence length is four. As in the case of length one gallery sets, we observe a monotonic improvement in the retrieval performance of all fusion methodologies as more images are fused. In summary, the proposed GRU based fusion techniques with and without m-loss significantly outperform the baseline fusion approaches and the effect is pronounced with increasing query sequence lengths, especially, when m-loss is additionally imposed while training the GRU.

\begin{figure}[t]
    \centering
    \begin{minipage}{0.8\linewidth}
        \begin{minipage}{0.49\textwidth}
            \centering
            \includegraphics[scale=0.24]{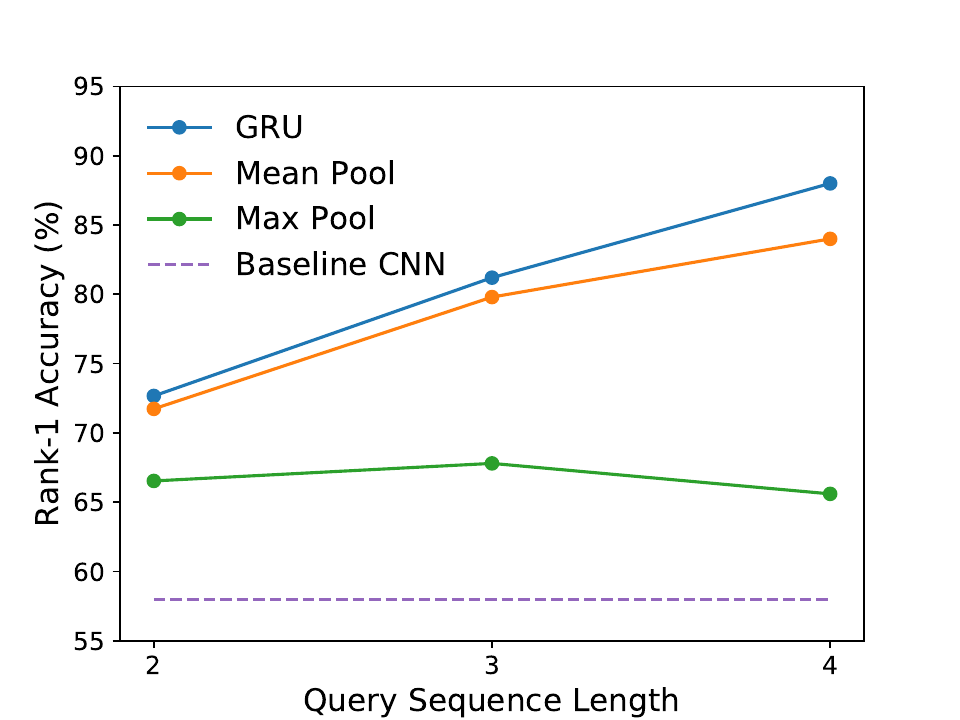}
        \end{minipage}
        \begin{minipage}{0.49\textwidth}
            \centering
            \includegraphics[scale=0.24]{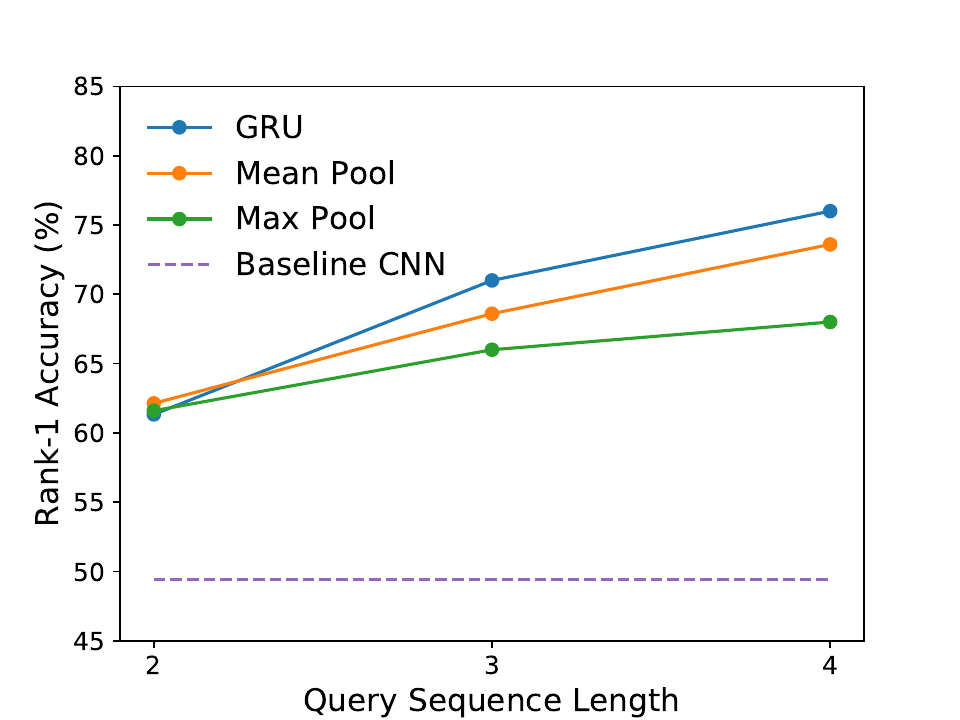}
        \end{minipage}
    \end{minipage}
    \caption{Rank-1 accuracy for Fixed Set Protocol on DukeMTMC-reID dataset with ResNet-50 (left) and AlexNet (right) CNN baselines.} 
    \label{fig:prot1_duke}
\end{figure}

Fig.~\ref{fig:prot1_duke} presents rank-1 accuracy results on the DukeMTMC-reID dataset following the FSP protocol. Due to dearth of query sequences with length greater than four, we consider query sets with a maximum of four cameras, while gallery size is fixed to two. The results are averaged over all such possible gallery sets. Our approach consistently outperforms other fusion techniques on both ResNet-50 and AlexNet baselines, while increasing the 
accuracy with fusion. We provide additional results on a third dataset (MSMT17~\cite{wei2018person}) in Section 4 of supplementary.

\begin{figure}[t]
    \begin{minipage}{1.0\linewidth}
    \flushleft
        \begin{minipage}{0.55\textwidth}
            \centering
            \includegraphics[width=1\linewidth]{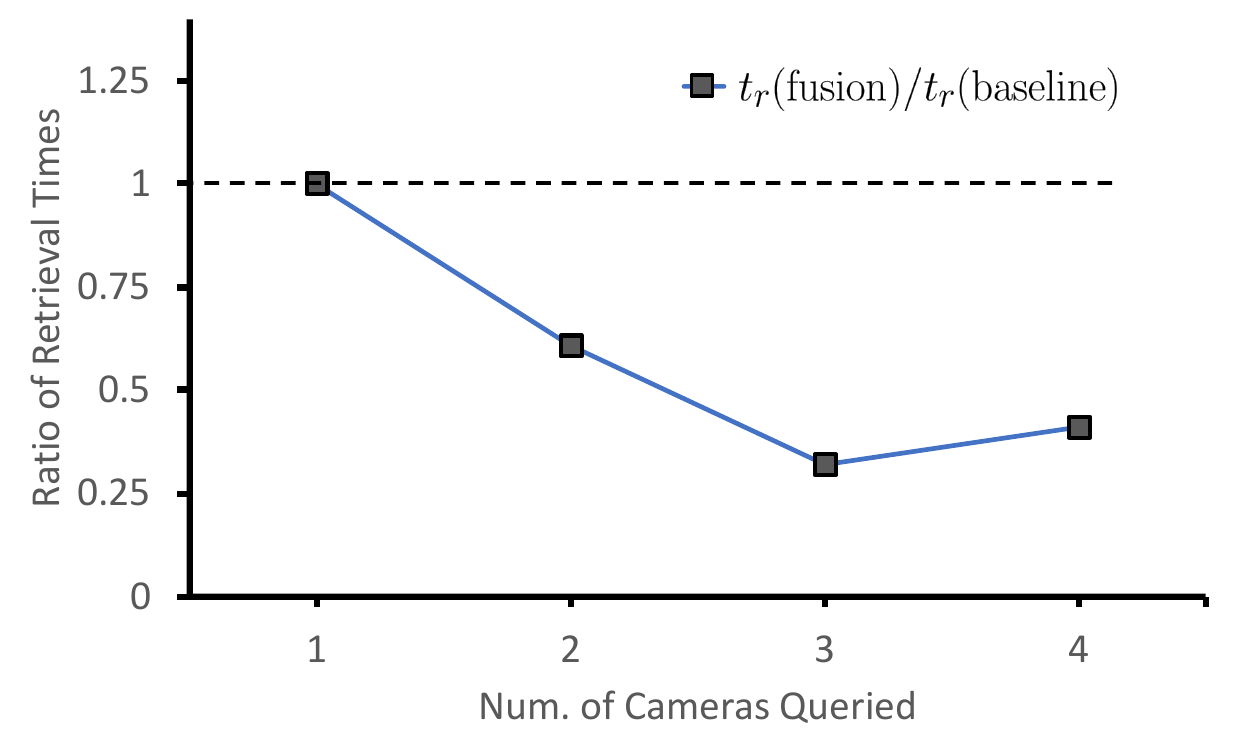}
        \end{minipage}
        \begin{minipage}{0.44\textwidth}
            \centering
            \includegraphics[width=1.0\linewidth]{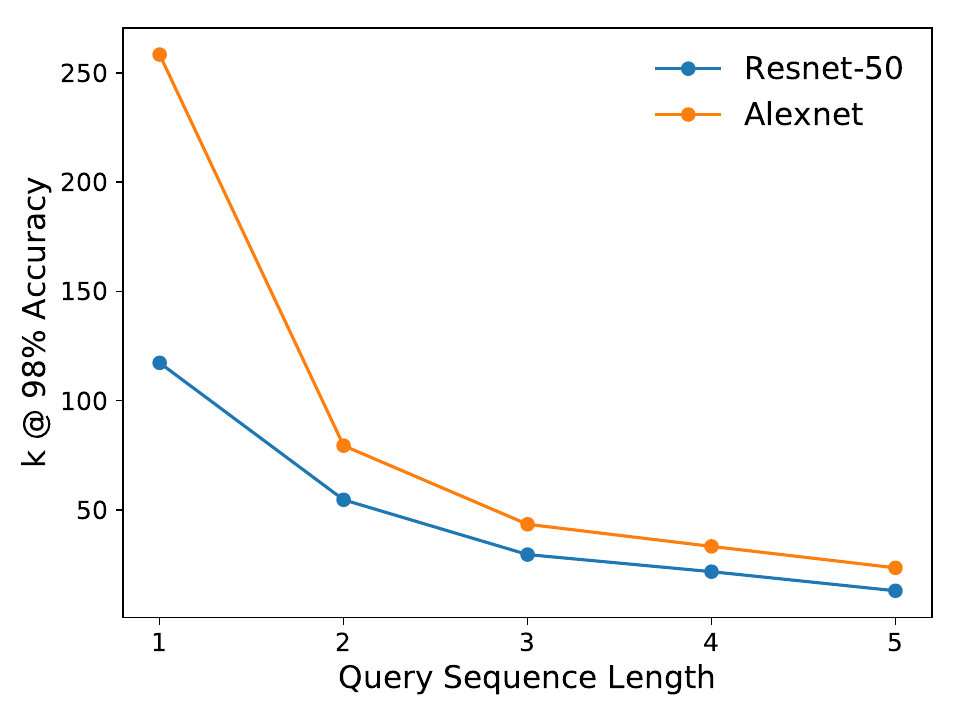}
        \end{minipage}
        \caption{Ratio of human operator retrieval times with and without fusion (left) and automated retrieval list length (right) plots on Market-1501 dataset.}
        \label{fig:time_plots}
    \end{minipage}
\end{figure}

\begin{figure*}[t]
    \centering
    \begin{minipage}{\linewidth}
        \begin{minipage}{0.49\textwidth}
            \centering
            \includegraphics[scale=0.23]{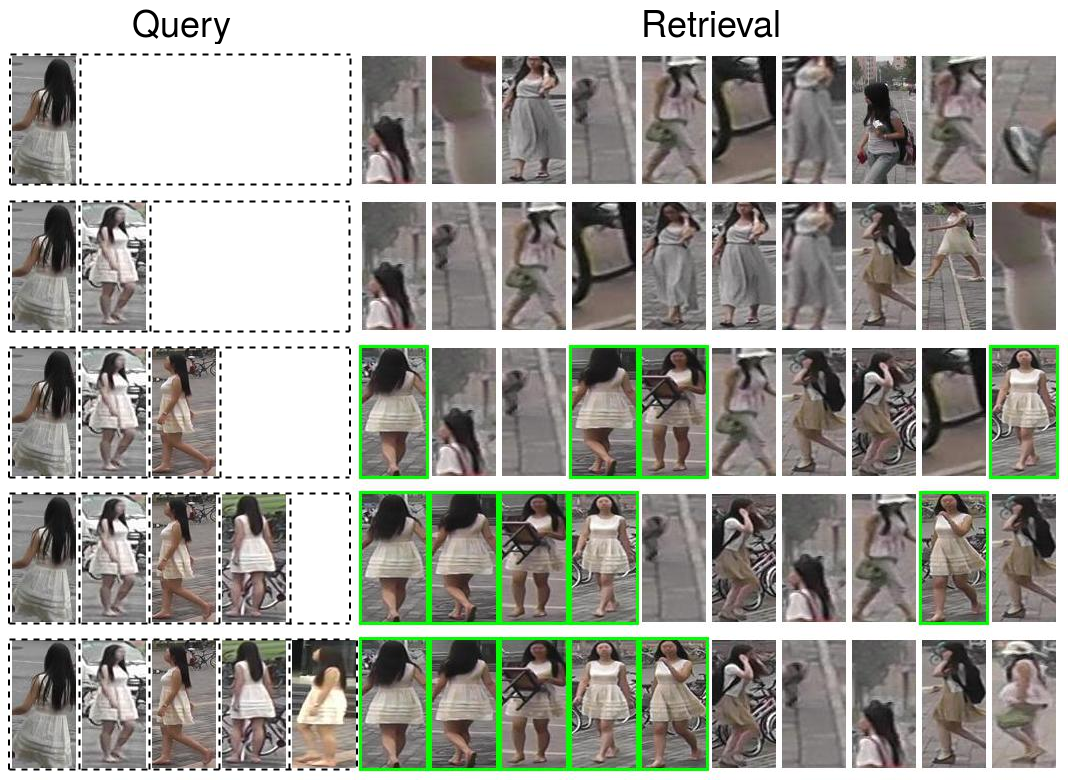}
        \end{minipage}
        \begin{minipage}{0.50\textwidth}
            \centering
            \includegraphics[scale=0.23]{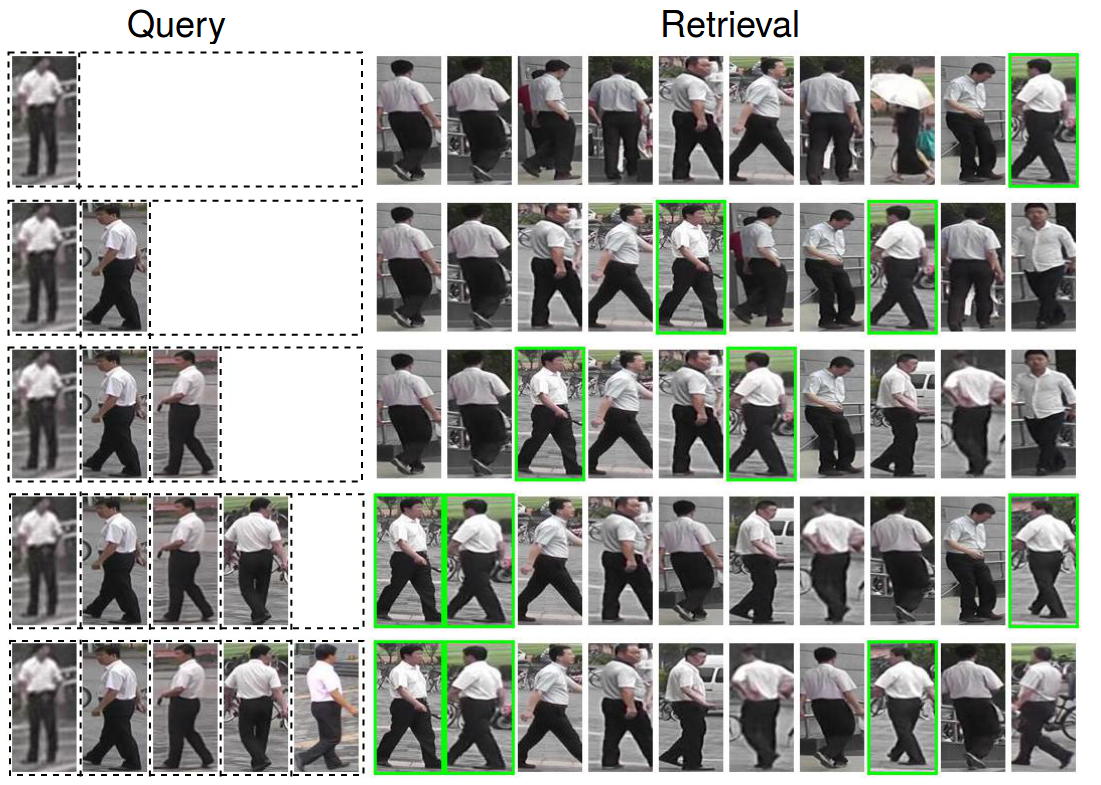}
        \end{minipage}
    \end{minipage}
    \caption{Retrieved samples for two example targets from Market-1501 dataset. Correct retrievals are indicated with green box. More correct matches are obtained at a lower rank as additional query images are combined (best viewed in color).}
    \label{fig:eg_vis}
\end{figure*}

\begin{figure}[t]
    \centering
    \begin{minipage}{0.9\linewidth}
        \begin{minipage}{0.49\textwidth}
            \centering
            \includegraphics[width=\linewidth]{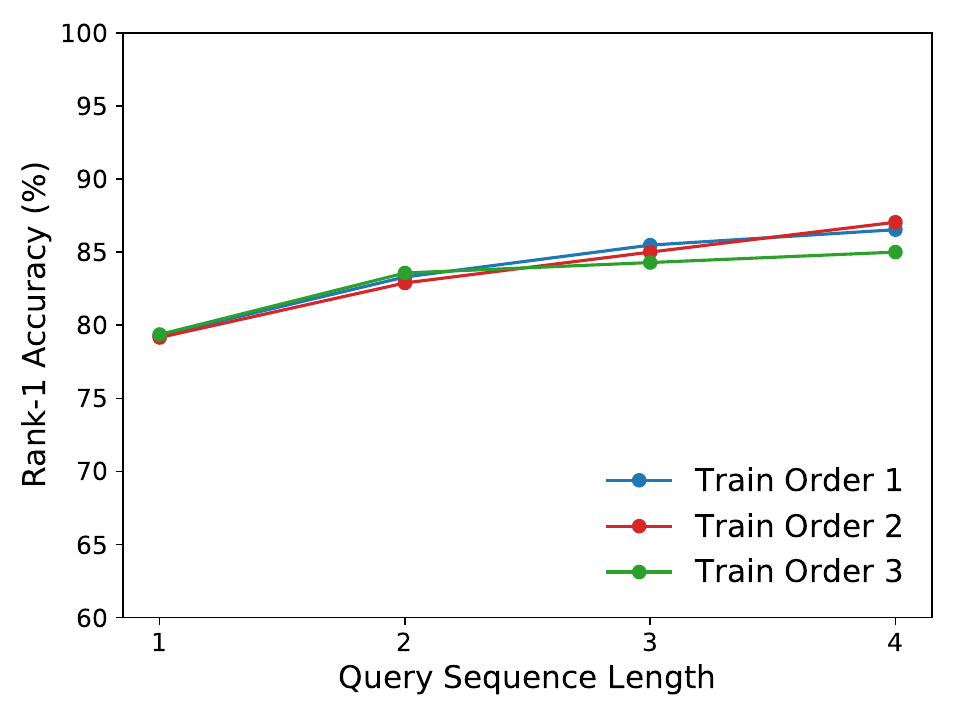}
        \end{minipage}
        \begin{minipage}{0.49\textwidth}
            \centering
            \includegraphics[width=\linewidth]{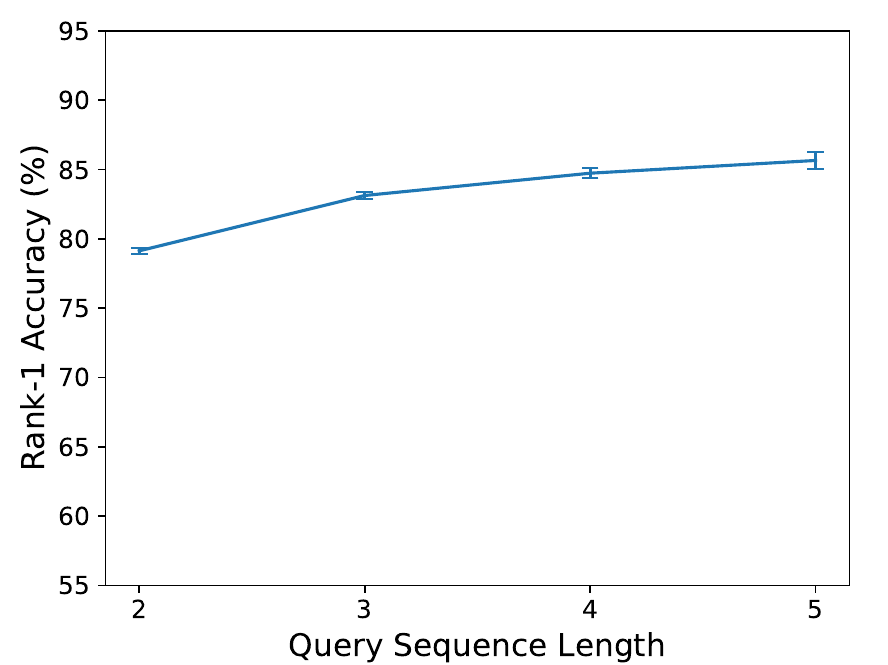}
        \end{minipage}
    \end{minipage}
    \caption{Averaged FSP results for different input ordering sequences during training (left) and testing (right).} 
    \label{fig:inputorder}
\end{figure}

\subsubsection{Comparison with Late Fusion Baselines}
\label{sec:latefusion}
In sections~\ref{subsec:vsp_results} and ~\ref{subsec:fsp_results}, we compared the proposed GRU based fusion with other early fusion schemes, namely mean and max pooling. We further substantiate our choice of fusion function through comparison with late fusion based approaches too.
Specifically, we use the features from the baseline CNN and perform score fusion and maximum probability based fusion. In both these schemes, fusion is done on the distances between query-gallery image features rather than at the feature level. That is, Euclidean distances between query and gallery image features are calculated independently for individual input features to be fused, which are subsequently combined using a weighted average to obtain the final distance values after fusion. In score fusion, equal weight is given to each of the input features to be aggregated. In maximum probability fusion, a discrete probability distribution over the gallery set is obtained by normalizing the distance of the query from the gallery images. The distance corresponding to the query input having maximum probability for a given gallery image is considered to be the fused distance. Quantitative results under both FSP and VSP protocols on the Market-1501 dataset are given in Table~\ref{tab:late_fusion}. In the case of FSP, we observe that the performance of the late fusion techniques are similar to that of other baseline mean pool based fusion scheme  and better than the max-pool scheme. In VSP, both the late fusion schemes are significantly better than mean/max pool for all query sequence lengths. However, the proposed GRU based fusion scheme consistently outperforms all the other baseline fusion schemes (early/late fusion) by a large margin.

\begin{table}[!t]
\resizebox{\linewidth}{!}
{%
    \begin{tabular}{|c|c|c|c|c|c|}
    \hline
    \multirow{2}{*}{Approach} &  \multirow{2}{*}{Protocol}& \multicolumn{4}{|c|}{Rank-1 @ Query Sequence Length}\\ \cline{3-6}
                              & & {2}     & {3}      & {4}       & {5}           \\ 
    \hline
    GRU+m-loss       & \multirow{3}{*}{FSP}              & \textbf{79.37}        & \textbf{83.58}           & \textbf{84.29}              & \textbf{85.01}  \\ 
    Score Fusion             &     & 73.60        & 77.30           & 79.11              & 80.44  \\
    Max Prob. Fusion         &     & 69.92        & 73.25           & 75.75              & 78.06  \\
    \hline
    GRU       & \multirow{3}{*}{VSP}         & \textbf{86.03}	    & \textbf{87.54}	    & \textbf{85.31}	    & \textbf{85.01}  \\ 
    Score Fusion             &     & 84.80	& 83.78	    & 81.17	    & 80.58  \\
    Max Prob. Fusion         &     & 80.96	& 79.50	    & 77.24	    & 78.19  \\
    \hline
    \end{tabular}
    }
\caption{Rank-1 accuracy comparison with late fusion approaches on Market-1501 dataset. Proposed GRU based fusion scheme consistently outperforms all the other baseline fusion schemes by a large margin.}
    \label{tab:late_fusion}
\end{table}

\subsubsection{Portability of the Fusion Scheme}
In the proposed fusion framework, the fusion function training is independent of the choice of feature extraction pipeline. The feature extraction network parameters are not updated during the training of fusion network. Though we choose a ResNet-50 based model trained on camera-pairwise re-identification tasks, the framework can easily accommodate any other general feature extraction pipeline. This plug-and-play nature of the proposed pipeline would enable us to seamlessly integrate any feature extractor that is used in traditional re-id setup, and the overall retrieval performance would surely benefit from any progress in the classical/traditional re-id. To further substantiate this claim, we show retrieval results of the proposed framework atop a state-of-the-art conventional person re-id approach. Specifically, we use the pre-trained model of HA-CNN~\cite{Li_2018_CVPR} to obtain the image feature representations as input to the GRU based fusion function. HA-CNN learns soft attention at the pixel level and hard attention at the region level and improves the feature representation through the use of a \lq{}harmonious attention\rq{} module. The training of our proposed fusion module with HA-CNN as feature extractor is done in a manner identical to that explained in the previous sections and the retrieval results are shown on Market-1501 dataset in Table~\ref{tab:hacnn}. We observe that the results are consistent with that obtained using the ResNet-50 baseline, i.e., the proposed re-id framework with the GRU fusion scheme achieving impressive improvements in retrieval performance over the baseline HA-CNN based re-id across both the FSP and VSP protocols. Also, it can be noted that the superiority of the fusion framework is apparent even in scenarios where the baseline achieves high retrieval accuracy.

\begin{table}[]
    \resizebox{1.\linewidth}{!}{
    \centering
    \begin{tabular}{|c|c|c|c|c|c|}
        \hline
    \multirow{2}{*}{Approach} & \multirow{2}{*}{Protocol}  & \multicolumn{4}{|c|}{Rank-1 @ Query Sequence Length}\\ \cline{3-6} 
                              & & {2}     & {3}      & {4}       & {5}           \\ 
    \hline
    GRU+m-loss       & \multirow{2}{*}{FSP}    & \textbf{83.65}	    & \textbf{85.69}	    & \textbf{86.13}	    & \textbf{87.50}           \\  \cline{3-6}
    HA-CNN Baseline       &     & \multicolumn{4}{|c|}{78.72}           \\  
    \hline
    GRU       & \multirow{2}{*}{VSP}    & \textbf{91.24}	    & \textbf{92.18}	    & \textbf{90.72}	    & \textbf{88.59}    \\ 
    HA-CNN Baseline      &              & 91.03	    &  89.06    & 85.63     & 83.58  \\ 
    \hline
    \end{tabular}
    }
    \caption{Portability of the proposed fusion framework: We observe that fusion improves the retrieval accuracy when used atop HA-CNN~\cite{Li_2018_CVPR}, a state-of-the-art re-id feature extraction pipeline.}
    \label{tab:hacnn}
\end{table}

\subsubsection{Advantages of Fusion in Deployed Systems}
To study the performance advantages of employing fusion-based algorithms in practical surveillance systems, we designed a prototype GUI system (Fig. 6 in supplementary materials) for human-operator-in-the-loop re-id and conducted a comparative user study to determine the relative time spent in retrieval with and without the fusion of queries. 

We showed $15$ different identities on an average to a pool of subjects recruited for the study. In the GUI, the query image is displayed on the left and the corresponding top-$k$ retrievals are displayed in the right panel in the order of increasing ranks (Fig. 6 in supplementary). We display $25$ retrievals ($k=25$) per page on the GUI. The subject searches through the retrievals and selects the matching image. If the subject is unable to find the right match, the next $k$ (25) retrievals are displayed. This process continues until the subject successfully locates a match. The retrieved image is then fused with the query to obtain retrievals in the subsequently queried camera. A similar experiment is performed without fusion, i.e., by querying each camera independently with one single image or retrieved target image from the preceding camera (without fusion). In Fig. 9 (left), we plot the ratio of the average time taken for retrieval with and without fusion ($t_r$(fusion)/$t_r$(baseline)) as a function of query sequence lengths (i.e., the number of cameras queried). We observe that retrieval times are significantly smaller and decrease with increasing query sequence length with our fusion-based approach in contrast to the conventional approach involving independent querying, thereby reinforcing the practical utility of the proposed framework.

The average rank of first correct retrieval (termed `minimum retrieval list length') as obtained by our algorithm is shown in Fig.~\ref{fig:time_plots} (right). The retrieval list length decreases monotonically with query sequence length, emphasizing the advantage of 
proposed approach. 

In Fig.~\ref{fig:eg_vis}, we present two sample sequences of queries and corresponding top-$10$ retrievals. 
As the fusion function processes more images, the number of correct retrievals within top-$10$ ranks increases. Fusion is especially beneficial in challenging scenarios where multiple candidates with near-identical appearances exist in the gallery with minute differences between them (Fig.~\ref{fig:eg_vis} (right)). Note that, while more correct retrievals are obtained within top-$10$ ranks as images are fused, there is an improvement in the position (rank) of the existing retrievals too. This indicates that our approach is able to integrate new information while retaining the relevant aspects of the existing representation. 

\subsubsection{Effect of Camera Ordering}
As discussed in Sec.~\ref{sec:proposedmethod}, we desire the fusion function to be agnostic to input ordering in both the training and testing phases. To verify this, we train the fusion network with multiple sequence orders corresponding to different camera arrangements. We observe that the average FSP results on six unit length galleries are similar across training orders (Fig.~\ref{fig:inputorder} (left)). Conversely, for a fixed training order, we examined multiple orderings of query cameras during testing. We sample 50 randomly ordered sequences of length $5$ and according to FSP (\ref{sec:fsp}), consider all possible combinations of sub-sequences for each sequence. The mean rank-1 accuracy and the standard deviations are plotted in Fig.~\ref{fig:inputorder} (right). As can be seen  (Fig.~\ref{fig:inputorder} (right)), the fusion performance is practically independent of camera ordering in this case as well. 

\section{Conclusion}
In this paper, we have proposed a novel sequential multi-camera feature fusion approach for person re-id. Unlike classical re-id methods, our approach can accommodate operator inputs in an online fashion, enabling early gains via a monotonic improvement in target retrieval accuracy. These capabilities are made possible by our choice of GRU as a fusion function and our training strategy involving a custom formulation of the monotonicity loss. We also introduce novel evaluation protocols and conduct extensive evaluations on Market-1501 and DukeMTMC-reID datasets. The results indicate that our multi-camera fusion method significantly outperforms the corresponding baselines as well as other popular feature fusion schemes. Additionally, our comparative analysis of operator-in-the-loop performance showcases the potential for seamless integration into deployable video-surveillance systems. 

Zheng et al.~\cite{zheng2018measuring} proposed a temporal metric for evaluation of re-id systems in a temporally changing dynamic gallery set scenario. It would be interesting to examine the connections between the temporal metric of Zheng et al. and the VSP protocol proposed in our current work since both deal with variable gallery sets. 
The current version of our work is not designed to explicitly omit noisy/spurious features from a camera, especially during the testing phase. One possibility would be to incorporate attention mechanisms in future to accomplish the same and further improve fusion during both training and testing phases.

\vspace{-5pt}
\section{Acknowledgement}
This work is partially supported by Pratiksha Trust, Bangalore and Robert Bosch Centre for Cyber Physical Systems, IISc.
\vspace{-10pt}
\bibliography{bibliography}
\bibliographystyle{IEEEtran}
\vskip -2\baselineskip plus -1fil
\begin{IEEEbiography}[{\includegraphics[width=1in,height=1.25in,clip,keepaspectratio]{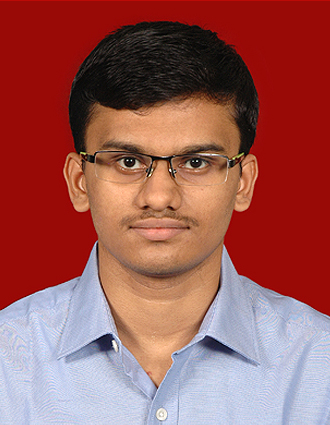}}]{K.L. Navaneet} is a masters student in Department of Computational and Data Sciences at Indian Institute of Science (IISc), Bangalore, India. He received his bachelors degree from Department of Electrical and Electronics Engineering, National Institute of Technology-Karnataka, Surathkal, India. His research interests include computer vision, machine learning and image processing.
\end{IEEEbiography}
\vskip -2\baselineskip plus -1fil
\begin{IEEEbiography}[{\includegraphics[width=1in,height=1.25in,clip,keepaspectratio]{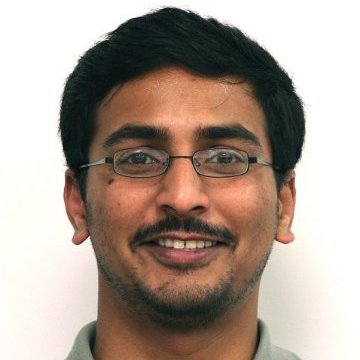}}]{Ravi Kiran Sarvadevabhatla} is  an  Assistant Professor at International Institute of Information Technology-Hyderabad (IIIT-H). He received his Ph.D. from Dept. of Computational and Data Sciences, Indian Institute of Science, Bangalore, India. He has broad-ranging research interests and likes to work on problems involving multi-modal multimedia data (e.g. images, videos, text, audio/speech, eye-tracking data) and multiple disciplines (e.g. Humanities, Graphics, Robotics, Human-Computer Interaction).
\end{IEEEbiography}
\vskip -2\baselineskip plus -1fil
\begin{IEEEbiography}[{\includegraphics[width=1in,height=1.25in,clip,keepaspectratio]{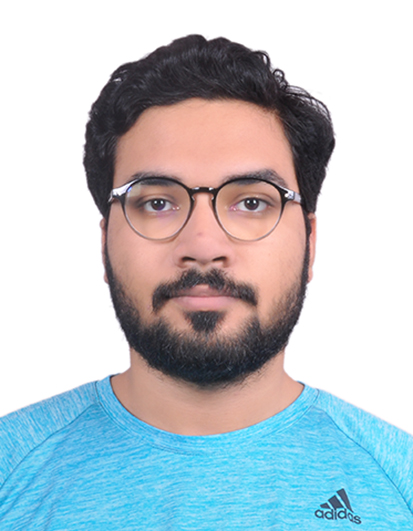}}]{Shashank Shekhar} is a Project Assistant at Visual Computing Lab, Department of Computational And Data Sciences, Indian Institute Of Science, Bangalore, India. He received his B.Tech. in Electronics And Communication Engineering from Indian Institute of Technology (ISM) Dhanbad, India in 2017 before working as a Software Engineer at Samsung Research Institute, Delhi for a while. His research interests include computer vision, representation learning and reinforcement learning.
\end{IEEEbiography}
\vskip -2\baselineskip plus -1fil
\begin{IEEEbiography}
    [{\includegraphics[width=1in,height=1.25in,clip,keepaspectratio]{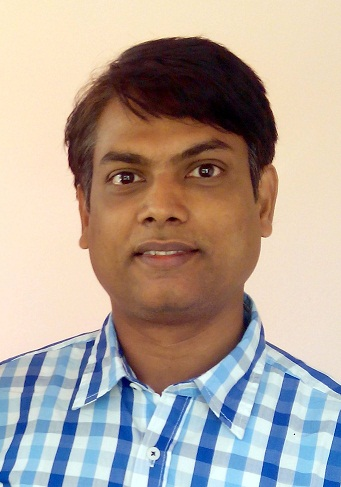}}]
    {R. Venkatesh Babu} is an Associate Professor in Department of Computational and Data Sciences, Indian Institute of Science, Bangalore. He received his Ph.D. degree from the Dept. of Electrical Engineering, Indian Institute of Science. Thereafter, he held postdoctoral positions at NTNU, Norway and IRISA/INRIA, Rennes, France. Subsequently he worked as a research fellow at NTU, Singapore. His research interests span signal processing, compression, machine vision, image/video processing, pattern recognition and multimedia. He is a senior member of IEEE.  
\end{IEEEbiography}
\vskip -2\baselineskip plus -1fil
\begin{IEEEbiography}[{\includegraphics[width=1in,height=1.25in,clip,keepaspectratio]{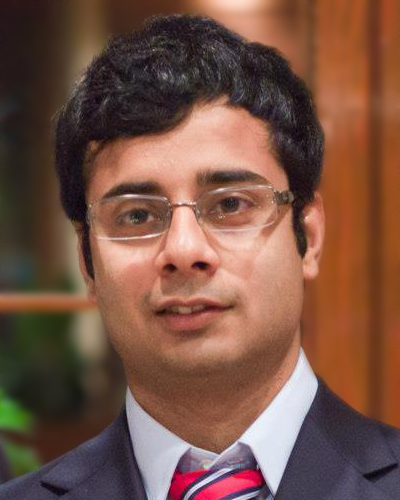}}]{Anirban Chakraborty} is an Assistant Professor in the Department of Computational and Data Sciences, Indian Institute of Science, Bangalore. He received his Ph.D. in Electrical Engineering from the University of California, Riverside in 2014. Subsequently, he held research fellow positions with the National University of Singapore and Nanyang Technological University. Anirban's research interests lie in the broad areas of computer vision, machine learning, optimization etc. and their applications in problems such as data association over large graphs, data fusion, video surveillance, video-based biometrics, bio-image informatics etc. He is a member of IEEE.
\end{IEEEbiography}
\vfill{}
\end{document}


\title{Operator-in-the-Loop Deep Sequential Multi-camera Feature Fusion for Person Re-identification - Supplementary Materials}

\author{K.~L.~Navaneet,
        Ravi~Kiran~Sarvadevabhatla,~\IEEEmembership{Member,~IEEE},
        Shashank Shekhar,
        R.~Venkatesh~Babu,~\IEEEmembership{Senior Member,~IEEE},
        and~Anirban~Chakraborty$^*$\thanks{* Corresponding author},~\IEEEmembership{Member,~IEEE}
\thanks{The authors are with the Department of Computational and Data Sciences, Indian Institute of Science, Bangalore, India, 560012. \break E-mail: navaneetl@iisc.ac.in, ravika@gmail.com, shashankshek@iisc.ac.in, venky@iisc.ac.in, anirban@iisc.ac.in}}


\maketitle

\IEEEdisplaynontitleabstractindextext

\IEEEpeerreviewmaketitle

\section{Introduction}
The supplementary material is organized as follows. We present the averaged retrieval results for FSP protocol over all gallery sets in the Market-1501 dataset (Sec~\ref{sec:fsp_market_all}). In Sec.~\ref{sec:vsp_duke} and ~\ref{sec:msmt}, we report additional quantitative results on DukeMTMC-reID and the large MSMT17 dataset. To analyse the effect of the relative weight for
m-loss, we provide results on the sensitivity of the retrieval performance to the weight parameter $\lambda$ in Sec~\ref{sec:lambda_ablation}. We provide more quantitative analysis of robustness of the learned representations on fusion in Sec.~\ref{sec:holistic}. Qualitative results with the deployment protocol is presented in Sec.~\ref{sec:qual_deployment}. In Sec.~\ref{computational_complexity}, we discuss the computational complexity aspects of the proposed sequential re-id framework. Finally, representative images of our GUI software are presented in Sec.~\ref{sec:gui}.

\section{Retrieval Results for FSP on Market-1501}
\label{sec:fsp_market_all}
\begin{figure}[t]
    \begin{center}
        \includegraphics[width=\linewidth]{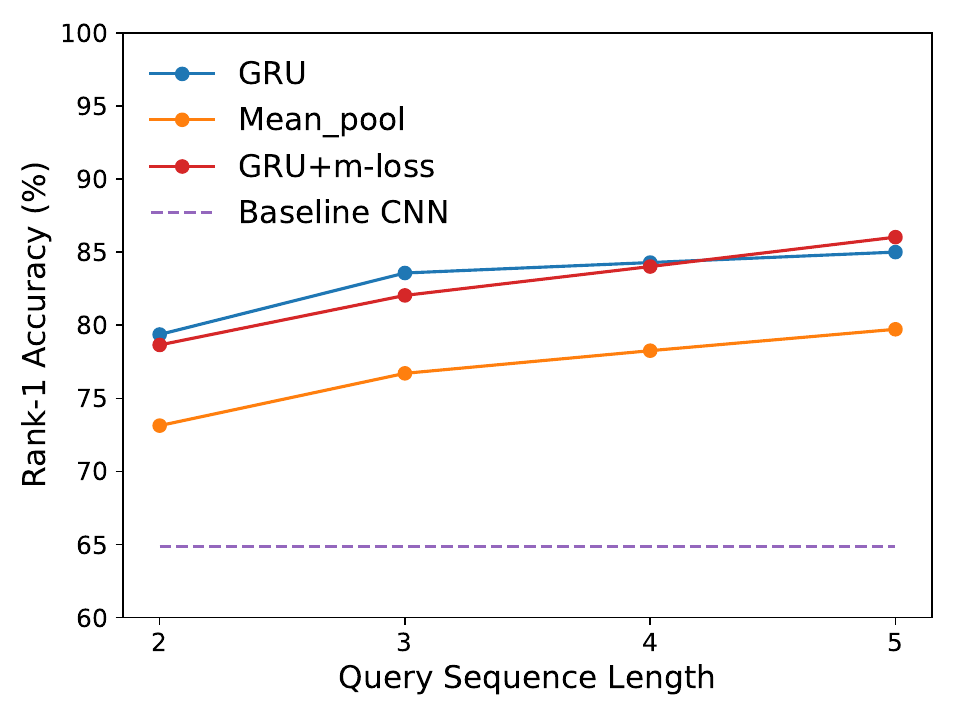}
    \caption{Effect of query sequence lengths on accuracy. Averaged Rank-1 accuracies are shown for fusion with ResNet-50 baseline on Market-1501 dataset.
    Feature fusion using GRU performs significantly better than other fusion techniques in rank-1 accuracies, while m-loss helps in maintaining monotonicity with increasing sequence lengths.}
    \label{fig:prot1_all}
\end{center}
\end{figure}

Fig. 6 of the paper presents FSP results on Market-1501 dataset for $3$ different gallery sets. Here, we present the averaged retrieval accuracy results for FSP protocol over all six galleries (Fig.~\ref{fig:prot1_all}). 

\section{Retrieval Results for VSP on DukeMTMC-reID}
\label{sec:vsp_duke}
\begin{figure}
    \centering
    \includegraphics[width=\linewidth]{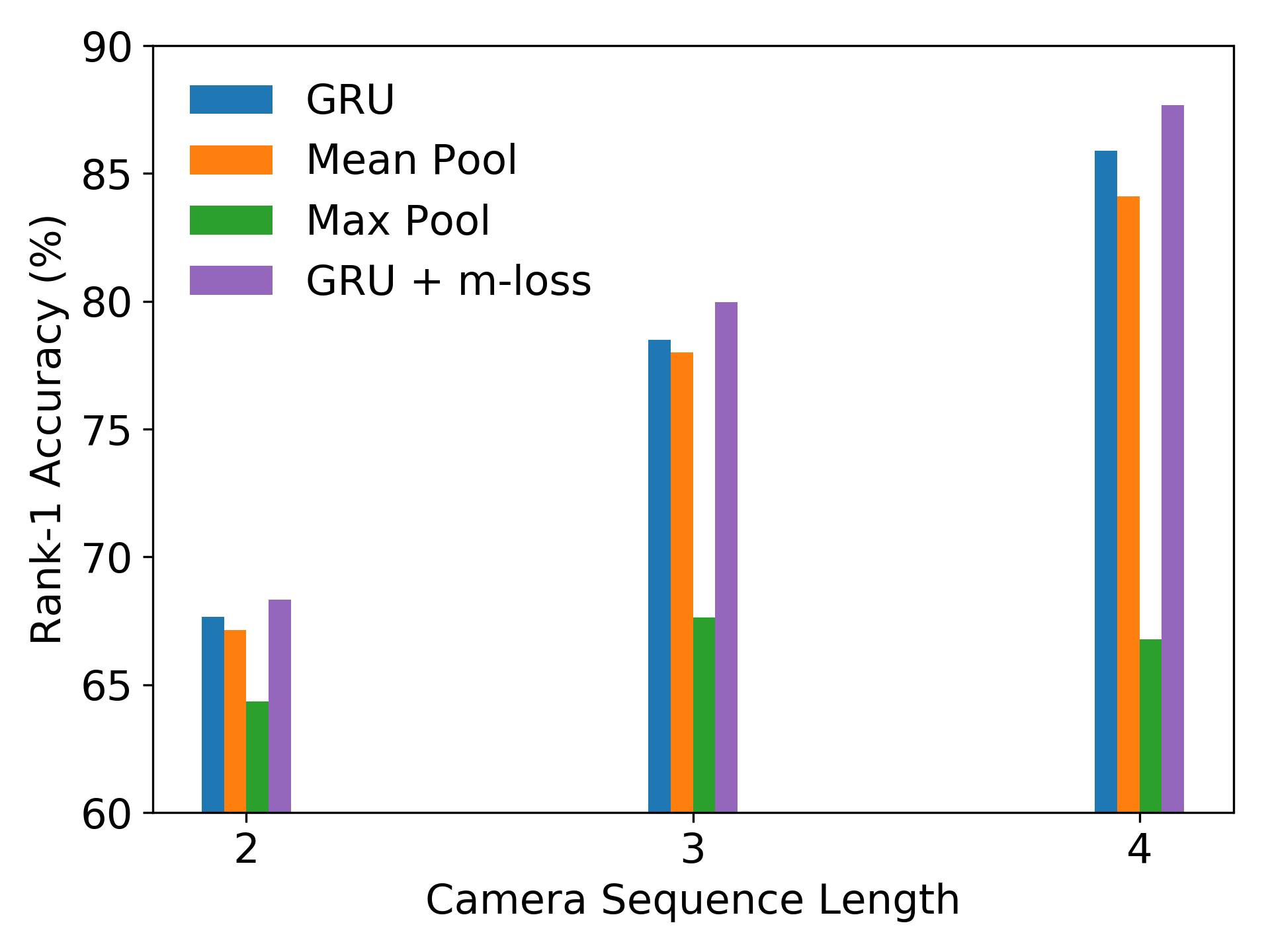}
    \caption{Retrieval results for VSP protocol on DukeMTMC-reID dataset. The proposed GRU based fusion function significantly 
outperforms the baseline fusion functions while the addition of m-loss yields performance improvements at longer query sequence lengths.}
    \label{fig:vsp_duke}
\end{figure}

Fig. 8 of the paper presents FSP results on DukeMTMC-reID dataset. Here, we present the retrieval accuracy results for VSP protocol for query sequence lengths 2 through 4. Note that since the number of common IDs for a given query-gallery camera subset might be very low, we only consider those combinations which have a minimum of 10 common identities. The results in Fig. ~\ref{fig:vsp_duke} demonstrate that the proposed GRU based fusion function significantly 
outperforms the baseline fusion functions while the addition of m-loss yields performance improvements at longer query sequence lengths. 

\section{Retrieval Results on MSMT17}
\label{sec:msmt}

\begin{figure}[]
    \begin{center}
        \includegraphics[width=\linewidth]{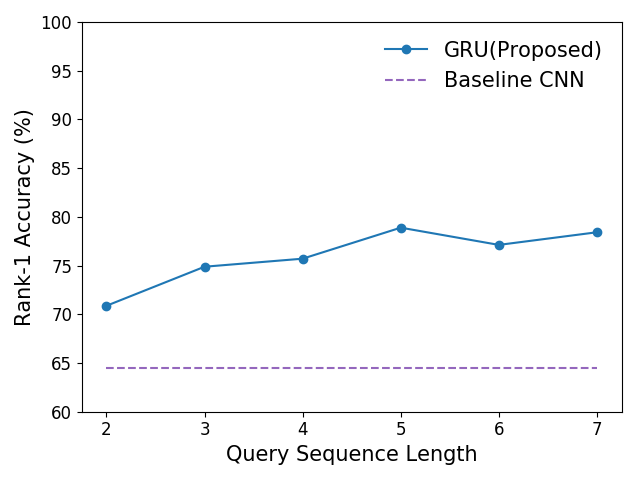}
    \caption{Retrieval results for FSP protocol on MSMT17 dataset. Comparison of averaged Rank-1 accuracies between fusion and baseline approach is presented. The proposed fusion framework achieves impressive improvement over the baseline in retrieval accuracy. The retrieval accuracy of the proposed fusion is near monotonic.}
    \label{fig:prot1_msmt}
\end{center}
\end{figure}

MSMT17~\cite{wei2018person} is the largest existing person re-identification dataset with $126,441$ bounding boxes of $4101$ unique identities captured in a network of $15$ cameras. The dataset is also challenging with images captured in both indoor and outdoor scenes and at different times of the day resulting in large illumination changes across the dataset. The large scale of the dataset enables effective training of deep neural network based re-id approaches. 
Fig. 6 and 8 of the paper present results on Market1501 and DukeMTMC-ReID, each containing images from $6$ and $8$ cameras respectively. Though MSMT17 dataset contains images from $15$ cameras, only one subset of $8$ cameras has at least $100$ common person IDs, a bare minimum for effective metric calculation. Thus, the maximum fusion sequence length is limited to just $7$, two more than Market-1501 dataset. However, as suggested by the reviewer, we present retrieval results for FSP protocol on MSMT17 dataset (Fig.~\ref{fig:prot1_msmt}). 
We observe that, similar to the case of Market-1501 (Fig. 6 of main paper) and DukeMTMC-reID (Fig. 8 of main paper), the proposed fusion framework significantly outperforms the baseline. The improvements in retrieval accuracy obtained using the proposed fusion is near monotonic.

\section{Sensitivity to Relative Weight of m-loss}
\label{sec:lambda_ablation}
The proposed loss formulation is a combination of two loss functions, namely triplet loss and monotonicity loss (m-loss) (refer to Eq. 6 of main paper). Triplet loss is aimed towards bringing features of same identities closer while pushing apart the features of different identities. In addition to the triplet loss, the m-loss is used to improve the feature representations at every step of feature fusion. m-loss tries to enforce sustenance or improvement of feature representations as more images are aggregated. However, a monotonic improvement in performance may be achieved at the cost of degraded accuracy. Thus the hyperparameters $\lambda$ and $\lambda_t^{R}$ are used to control the relative importance of the losses in the overall optimization. We perform an empirical analysis to decide the optimal value of $\lambda$ (Table~\ref{tab:lambda_acc}). While a very low value of lambda ($\lambda<0.0001$) results in a performance similar to that without m-loss where monotonicity is not enforced, very high values of lambda($\lambda>1$) result in stricter enforcement of monotonicity at the cost of retrieval performance. The retrieval accuracies are not very sensitive to lambda in the range ($0.001, 10$). Based on the empirical study, we set the value of lambda to $0.01$. \newline

\begin{table}[]
    \centering
    \begin{tabular}{cccccc}
        \toprule
        $\lambda$ & 0.001 & 0.01 & 0.1 & 1.0 & 10  \\
        Rank-1 Acc (\%) & 79.15 & 80.13 & 77.49 & 79.72 & 78.33 \\
        \bottomrule
    \end{tabular}
    \caption{Effect of $\lambda$ on the retrieval accuracy. The averaged rank-1 accuracies for FSP are reported for varying values of $\lambda$. The retrieval performance is consistent over a wide range of values for $\lambda$ and the 
             best performance is achieved for $\lambda=0.01$.}
    \label{tab:lambda_acc}
\end{table}

\section{Robustness of Feature Representations}
\label{sec:holistic}

\begin{figure}[]
    \begin{center}
        \includegraphics[width=\linewidth]{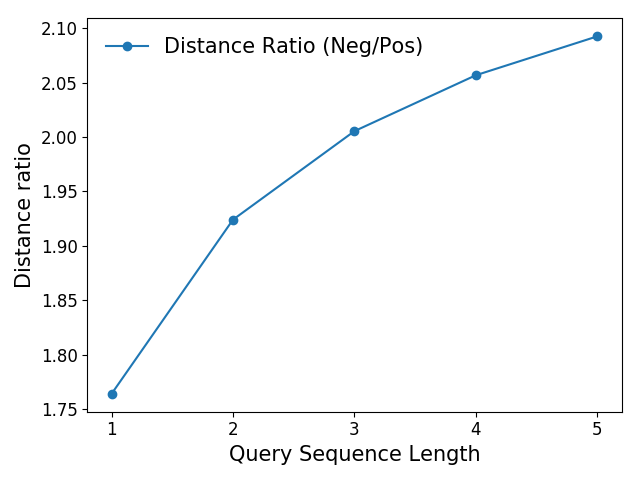}
    \caption{Robustness of feature representations. We compute the ratio of Euclidean distances between negative (different ID) and positive (same ID) samples at different fusion sequence lengths. Length 1 indicates no fusion. The plot indicates that the fusion results in improved feature representations. We also observe that the improvement is monotonic with the number of cameras fused.}
    \label{fig:holistic}
\end{center}
\end{figure}

In the proposed fusion framework, a camera network is sequentially queried and human operator feedback in the form of retrieved images are fused with existing inputs at the feature level to further query the remaining cameras in the network. An additional loss termed m-loss is employed to ensure that the retrieval performance is either improved or sustained as more inputs from different cameras are aggregated. Fig. 6 and 8 in the paper present retrieval accuracy on the Market-1501 and DukeMTMC-reid respectively. We observe that monotonic increase in rank-1 accuracy as the query sequence length increases, indicating that fusion at every step results in an improved feature representation. 
In Fig. 5 of main paper, we show a comparative quantitative analysis of different fusion schemes with the VSP protocol for varying query sequence lengths. Note that as the query sequence length, that is, the number of cameras in the query subsets increases, the number of cameras in the gallery subsets (which is the complementary set of query subset) decreases. Thus the number of correct matches in the gallery set decreases for higher query sequence length. We observe that this results in a significant drop in rank-1 accuracy for the baseline fusion approaches (mean/max pool) as more images are fused. However, the proposed GRU based fusion scheme maintains a high retrieval performance even when the number of gallery cameras are reduced. This demonstrates that the proposed scheme results in a more effective and robust fusion. 
To further demonstrate that the proposed scheme generates a more robust and holistic feature representation, we provide an additional quantitative analysis on the fused feature representations (Fig.~\ref{fig:holistic}). Specifically, we compute the Euclidean distance between every query and gallery feature representations. We calculate \lq{}$d_{pos}$\rq{} and \lq{}$d_{neg}$\rq{}, the average distance of each query with all the positive (images of the same person ID as the query) and negative (images corresponding to ID different from that of query) samples in the gallery set. Higher the $d_{neg}$, greater is the separation between feature representations of different classes (PIDs). Similarly, lower the $d_{pos}$, closer the features belonging to the same class. The ratio of $d_{neg}$ to $d_{pos}$ is presented in Fig.~\ref{fig:holistic}. A higher ratio is indicative of greater inter-class and lower intra-class distances, and thus of more robust and holistic feature representations. We observe that the ratio increases monotonically, demonstrating that the feature representations achieve improvement at every stage of fusion.

\section{Qualitative Results for Deployment Framework}
\label{sec:qual_deployment}

\begin{figure*}[]
    \begin{center}
        \includegraphics[width=\linewidth]{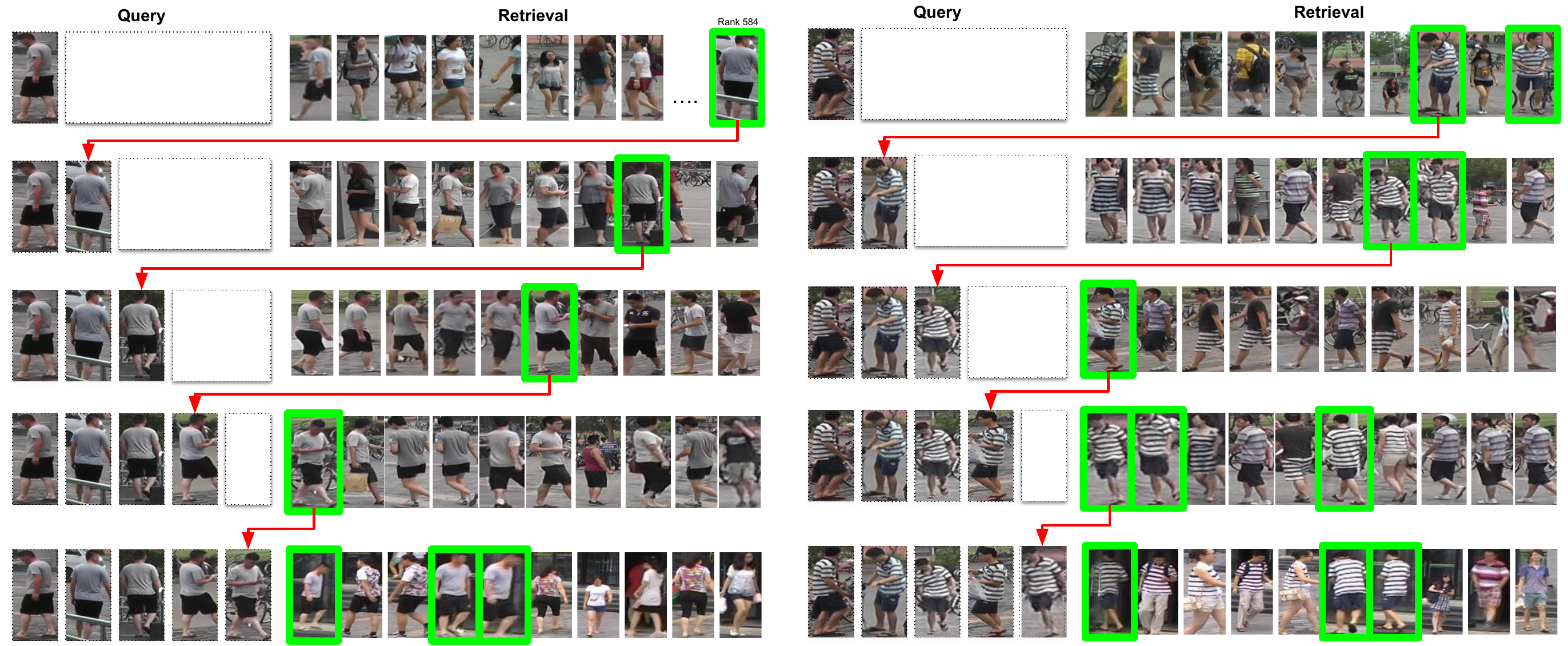}
    \caption{Qualitative results with the deployment framework. A gradual improvement in the representation for the query target is achieved as retrieved images from the gallery cameras are iteratively fused. This improvement is evident by the rapid increase in the low rank retrieval accuracy.}
    \label{fig:qualitative_deployment}
\end{center}
\end{figure*}

Fig. 10 of the main paper presents qualitative results with the FSP protocol described in Sec 4.3. The FSP protocol is, in principle, similar to the deployment scenario depicted in Fig. 1 of the main paper. However, it is suitably modified to enable effective quantitative analysis of the proposed fusion functions. Thus, while it is not possible to use the deployment framework for quantitative evaluation of fusion functions, here we present qualitative results of the same (Fig.~\ref{fig:qualitative_deployment}). The figure presents query and corresponding top-$10$ retrievals from the gallery camera for different fusion sequence lengths. The retrieved image is fused with the query at the feature level to query the subsequent camera. Similar to Fig. 10, it can be observed that a gradual improvement in the representation for the query target is achieved as retrieved images from the gallery cameras are iteratively fused. This improvement is evident by the rapid increase in the low rank retrieval accuracy, as shown in the figure.

\section{Computational Complexity of Proposed Framework}
\label{computational_complexity}
The overall pipeline of the proposed sequential re-id scheme consists of two main components. To obtain a list of top retrievals, we first use any off-the-shelf CNN feature extraction module, followed by a recurrent fusion function that combines these features extracted from the same target across multiple cameras. Thus, the worst-case complexity of the proposed overall framework would be $O(N)$, where $N$ is the total number of cameras in a network. However, due to the specific choice of our fusion function (GRU) as well as the on-demand nature of a typical re-id problem, it is possible to implement the framework in a more efficient manner. There are often large ``blind gaps" between different cameras, which may result in significant delays between the disappearances and re-appearances of a target across these camera FoVs. In view of this, as the target transits between its $(t-1)^{\text{th}}$ and $t^{\text{th}}$ appearances in a network, the fused feature (same as the hidden state representation $h_{t-1}$ of the GRU) can be computed and stored until the target re-appears for the $t^{\text{th}}$  time. Once the target is observed and identified by operator, the final fused representation can be computed by performing just a single step of update (i.e., $h_t = g(x_t, h_{t-1})$, where $g$ can be described using Eqns. 7a - d in the manuscript), thereby resulting in a constant complexity for our fusion function, for all practical purposes.

\section{GUI of Prototype Software}
\label{sec:gui}

We show images from our prototype deployment software for the sequential fusion framework. Query images are shown on the left, and top-k (in batches of 25 per page of the GUI) retrieved images are presented on the right. As each camera is queried, the retrieved and correctly identified image is fused with the existing inputs to query the subsequent cameras. 

\begin{figure*}[!t]
    \centering
    \includegraphics[width=0.95\linewidth]{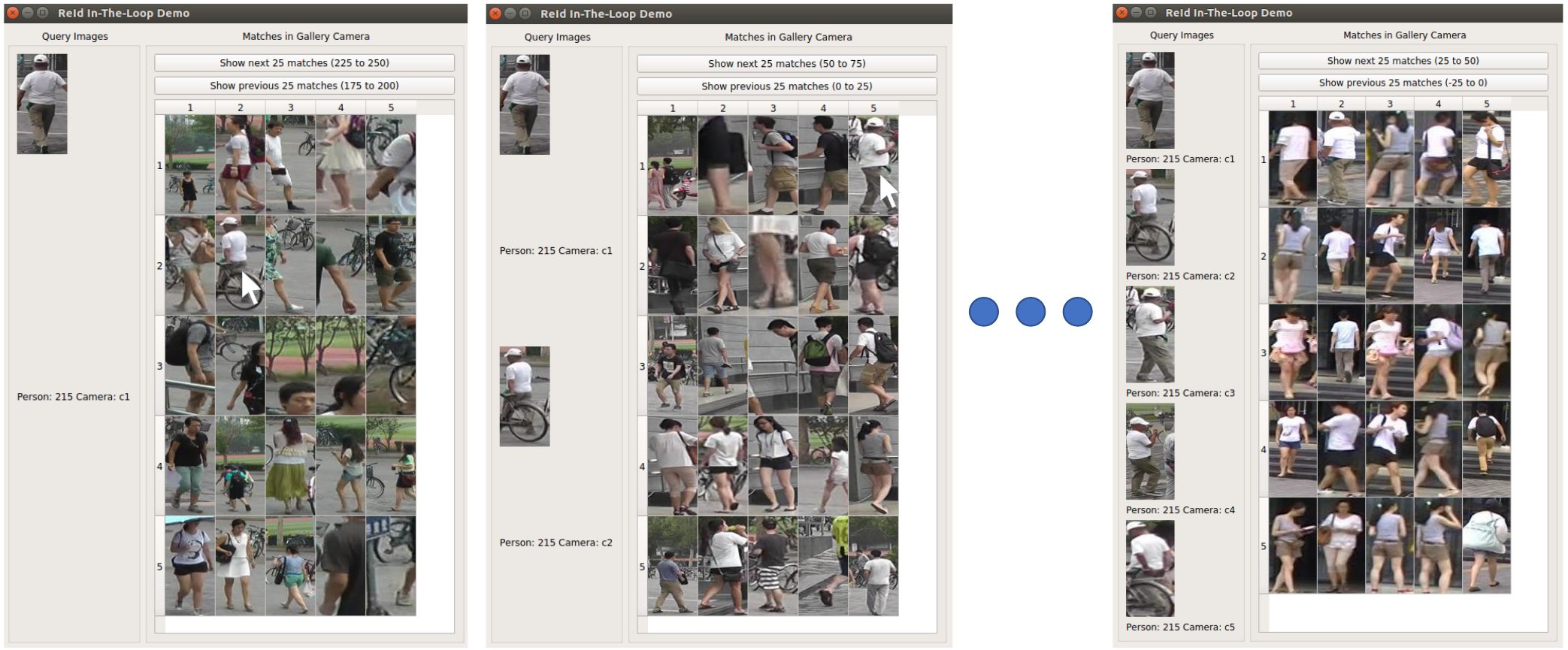}
    \caption{Sample interaction of operator with the prototype user-interface for our fusion based re-id system.}
    \label{fig:demo}
\end{figure*}

\bibliography{bibliography}
\bibliographystyle{IEEEtran}